\documentclass[journal]{IEEEtran}

\ifCLASSINFOpdf
\else
\fi
\usepackage{times}
\usepackage{xcolor}
\usepackage{soul}
\usepackage[utf8]{inputenc}
\usepackage[small]{caption}
\usepackage{multirow}
\usepackage{graphicx}  
\usepackage{helvet}
\usepackage{courier}
\usepackage{url}
\usepackage{color}
\usepackage{amsmath}
\usepackage{amssymb}
\usepackage{algorithm}
\usepackage{algorithmic}

\begin{document}
%
\title{Learning Dynamic Compact Memory Embedding for Deformable Visual Object Tracking}
%
\author{Pengfei Zhu, Hongtao Yu, Kaihua Zhang, Yu Wang, Shuai Zhao, Lei Wang, Tianzhu Zhang, Qinghua Hu
\IEEEcompsocitemizethanks{
\IEEEcompsocthanksitem This work was supported by the National Key Research and Development Program of China under Grant 2018AAA0102402, the National Natural Science Foundation of China under Grants 61732011, 61876127, 61925602, and 61876088, Natural Science Foundation of Tianjin under Grant 17JCZDJC30800, the Applied Basic Research Program of Qinghai(2019-ZJ-7017).
\IEEEcompsocthanksitem Pengfei Zhu, Hongtao Yu, Yu Wang, and Qinghua Hu are with the College of Intelligence and Computing, Tianjin University, Tianjin, China (e-mail: \{zhupengfei,yuhongtao,wangyu\_,huqinghua\}@tju.edu.cn). 

Kaihua Zhang is with School of Computer and Software, Nanjing University of Information Science and Technology, Nanjing, China (e-mail:zhkhua@gmail.com).

Tianzhu Zhang is with School of Information Science and Technology, University of Science and Technology of China (USTC), Hefei, China (Email: tzzhang@ustc.edu.cn).

Shuai Zhao is with the College of Intelligence and Computing, Tianjin University, Tianjin, China, and is also with the Automotive Data of China (Tianjin) Co., Ltd, Tianjin, China (e-mail:zhaoshuai@catarc.ac.cn).

Lei Wang is with the Automotive Data of China (Tianjin) Co., Ltd, Tianjin, China (e-mail:wanglei01@catarc.ac.cn).

}
}

\markboth{}%
{Zhu \MakeLowercase{\textit{et al.}}: Learning Dynamic Compact Memory Embedding for Deformable Visual Object Tracking}

\maketitle

\begin{abstract}
Recently, template-based trackers have become the leading tracking algorithms with promising performance in terms of efficiency and accuracy. However, the correlation operation between query feature and the given template only exploits accurate target localization, leading to state estimation error especially when the target suffers from severe deformable variations.
To address this issue, segmentation-based trackers have been proposed that employ per-pixel matching to improve the tracking performance of deformable objects effectively.
However, most of existing trackers only refer to the target features in the initial frame, thereby lacking the discriminative capacity to handle challenging factors, e.g., similar distractors, background clutter, appearance change, etc.
To this end, we propose a dynamic compact memory embedding to enhance the discrimination of the segmentation-based deformable visual tracking method.
Specifically, we initialize a memory embedding with the target features in the first frame. During the tracking process, the current target features that have high correlation with existing memory are updated to the memory embedding online. To further improve the segmentation accuracy for deformable objects, we employ a point-to-global matching strategy to measure the correlation between the pixel-wise query features and the whole template, so as to capture more detailed deformation information.
Extensive evaluations on six challenging tracking benchmarks including VOT2016, VOT2018, VOT2019,  GOT-10K, TrackingNet, and LaSOT demonstrate the superiority of our method over recent remarkable trackers. Besides, our method outperforms the excellent segmentation-based trackers, i.e., D3S and SiamMask on DAVIS2017 benchmark.

\end{abstract}


\section{Introduction}
\begin{figure}[htbp]
 	 \centering
	\includegraphics[width=\linewidth]{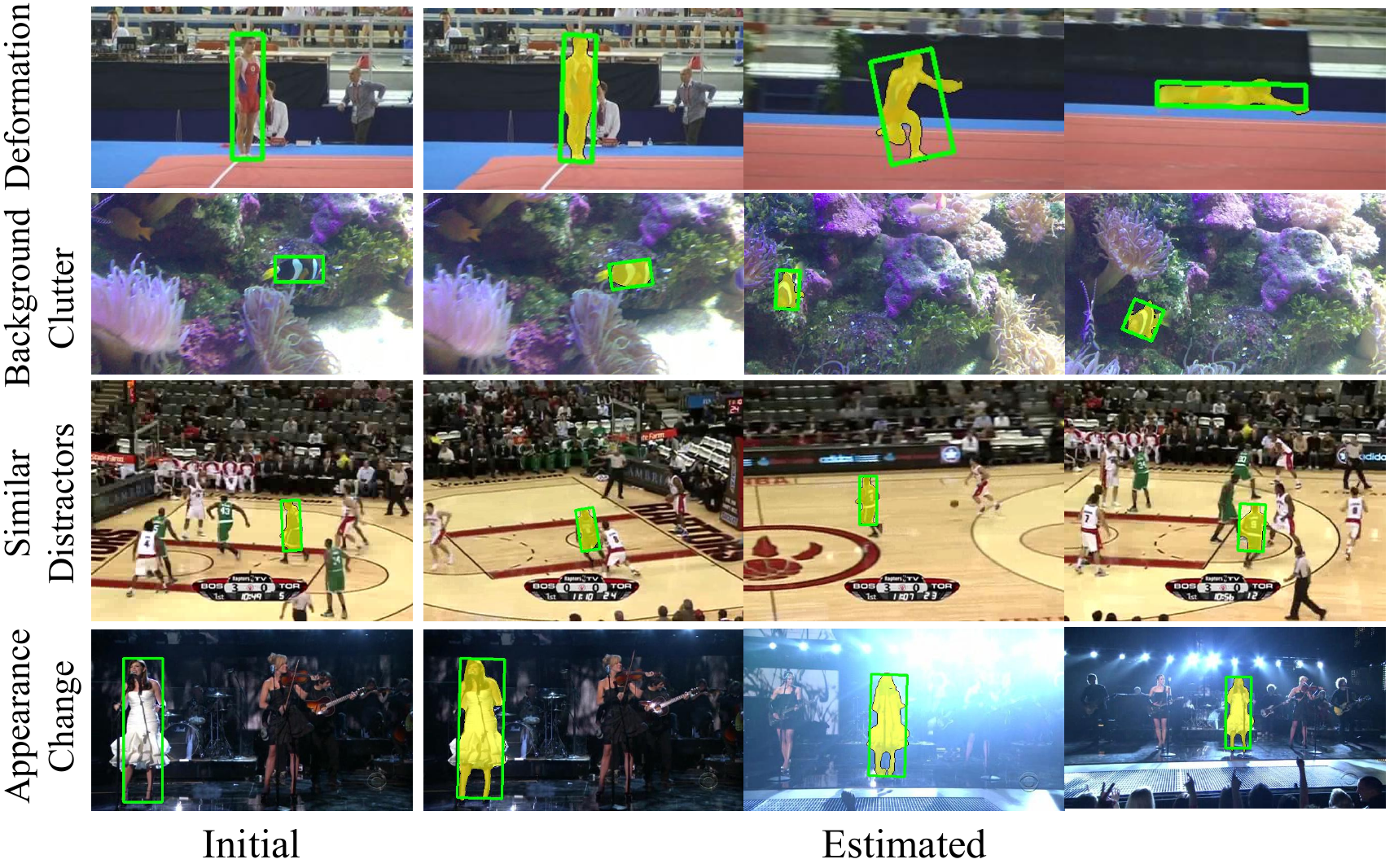}
	\caption{Some sampled qualitative results of our method on four challenging video sequences, i.e., \textit{gymnastics1}, \textit{fish4}, \textit{basketball}, and \textit{singer1}. Our method overcomes the challenges (e.g., object deformation, background clutter, occlusion, similar distractors,  appearance change, to name a few) in the above sequences and achieves remarkable tracking and segmentation performance.}
	\label{fig:introduction}
\end{figure}

Visual object tracking (VOT) is a fundamental and challenging task in computer vision community. It has numerous practical applications, such as traffic surveillance, human-computer interaction, autonomous robots, autonomous driving, etc.~\cite{Xing2010MultipleHT, Liu2012HandPR, Lee2015OnRoadPT, Tang2017MultiplePT}. Although VOT achieves great improvement in terms of both accuracy and robustness, yet, there are some remaining challenges needed to be solved, e.g., similar distractors, background clutter, deformation, etc.

In recent years, siamese network-based VOT methods have attracted widespread attention due to their high tracking speed and accuracy.
SiamFC \cite{Siamfc} and SINT \cite{SINT} are the pioneers of siamese network-based trackers.
They apply the strategy of multi-scale search to estimate the target scale state. 
Based on the original works, the developed anchor-based  \cite{SiamRPN, CascadedRPN, SiamRPN++, SiamDW, SPM} and anchor-free \cite{Siamfc++, SiamCAR, SiamBAN, Ocean} methods adopt target scale regression strategy, which effectively improves tracking performance for target scale variation.
However, it is hard for a fixed template to adapt to changable scenarios and target appearance, and it may lead to mismatches between the template and the search region. Especially for the situations with obvious target appearance change and similar obejcts, the siamese network-based trackers usually fail.
On the other line, the discriminative correlation filters (DCF) based trackers \cite{MOSSE, ECO, CCOT, Xutianyang, ATOM, ASRCF, LADCF, DiMP} employ ridge regression to learn the template, which is updated online. To a certain extent, DCF-based trackers can alleviate the dilemma that fixed templates are hard to adapt to scene changes.
Despite demonstrated success, siamese network-based and DCF-based tracking algorithms have a common drawback, that the limited template information and correlation-based matching method are unable to fully express the deformation of the target.
To solve this problem, a series of promising segmentation-based trackers are proposed~\cite{Siammask, D3S}. These trackers are combined with video object segmentation (VOS) model to exploit the insensitivity of segmentation methods to non-rigid deformations.
%
It is worth noting that the encoding feature inputted to the segmentation network is generated from target similarity matching between the current query feature and the template from the initial frame.
However, the template feature cannot cover all the expressions of the target to be tracked within the whole video sequence.
In addition, the result of target similarity matching usually contains a certain amount of mismatching noise caused by similar distractors, background clutter, occlusion, etc. Therefore, the matching result cannot express accurate spatial position and scale status of the target.
In summary, the template matching in VOT has the following two limitations. Firstly, a single template cannot provide enough target information for target similarity matching. The initial template only contains limited and fixed target structure information, which cannot adapt to target appearance variations over time in the video sequence. Secondly, the existing matching methods conduct between corresponding pixels merely, which is too rough. It is hard to capture fine target deformation sufficiently.

To address the above issues, in this paper, we propose to learn dynamic compact memory embedding (CME) for deformable visual tracking.
Inspired by the Hash algorithm~\cite{Monga2006PerceptualIH}, we develop a dynamic memory embedding mechanism for target similarity matching. By retrieving the feature affinity matrix which is  generated during target similarity matching process, we obtain the correlation between the query feature and the existing memory. Then, we merge the high-correlation parts between the existing memory and the current target feature. In this way, we form a memory without repetition. Besides, we concatenate the parts with medium-correlation to the memory, and directly discard the irrelevant ones. Therefore, the diversity and compactness of the memory embedding can be ensured. High-quality memory embedding provides the complete target information in historical frames, thus effectively dealing with target occlusion, similar distractors, and other variations. To further perceive the target deformation, we propose a deformable feature learning module (DFL). 
Compared with the existing methods based on correlation operation~\cite{Siamfc, SiamRPN++, SiamRPN} or target similarity matching~\cite{SiamRPN++},
our proposed deformable feature learning method adopts a pixel-to-global association strategy. By aggregating the weighted correlation between the each query pixel and the entire reference feature, the target deformation state of the target can be effectively captured. For simplicity, our model is abbreviated as CMEDFL.

The effectiveness of our model is verified on six visual tracking benchmarks: VOT2016 \cite{VOT16}, VOT2018 \cite{VOT18}, VOT2019 \cite{VOT19},  GOT-10K \cite{GOT10K}, TrackingNet \cite{TrackingNet}, and LaSOT\cite{LaSOT}.
Our CMEDFL obtains a new remarkable EAO score of 0.525 on VOT2018 \cite{VOT18}. Compared to recent state-of-the-art trackers such as Ocean \cite{Ocean}, D3S \cite{D3S}, DiMP \cite{DiMP}, etc.  our model achieves competitive performance. Fig.~\ref {fig:introduction} illustrates some sampled qualitative results of our method on four challenging video sequences. Our CMEDFL successfully overcomes the challenges of occlusion, deformation, background clutter, similar distractors, etc.

The main contributions of our work can be summarized as threefold :

$\left(1\right)$ We propose a CME for target similarity matching. The dynamic compact memory adjustment mechanism only stores high-quality historical information of the target, thus providing effective reference template in complex situations such as similar distractors or background clutter.

$\left(2\right)$ We propose deformable feature learning  to further improve the accuracy of segmenting deformable target. The deformation of the target can be effectively obtained by establishing the global correlation between the per-pixel query feature and the entire reference template feature.

$\left(3\right)$ Abundant comparison experiments conducted on six challenging tracking  datasets show that our CMEDFL achieves remarkable performance compared to several state-of-the-art trackers. Moreover, CMEDFL outperforms the state-of-the-art segmentation-based trackers (i.e., D3S \cite{D3S} and SiamMask \cite{Siammask}) and the general VOS methods (e.g., VM \cite{VideoMatch} and FAVOS \cite{FAVOS}) on DAVIS2017 \cite{DAVIS17} benchmark.

\section{Related Work}

\subsection{Visual Object Tracking}
\textbf{Siamese Network-based Tracking.} During tracking, the position of the maximum correlation between the search region and the fixed template is considered as the target localization. For better generalization, siamese network-based trackers are usually trained by massive labeled data. SINT \cite{SINT} and SiamFC \cite{Siamfc} have a milestone impact on the development of visual tracking. They are the first attempt to train the siamese networks end-to-end for visual tracking. SiamRPN++ \cite{SiamRPN++} and SiamDW \cite{SiamDW} improve the structure of ResNet \cite{ResNet} and successfully apply it to the siamese network-based tracking model, which significantly improves the tracking performance. SiamRPN \cite{SiamRPN} applies the region proposal network (RPN) to the siamese network for tracking. The two-branch network has the classification head for foreground-background separation of anchors, and the regression head for proposal refinement. Compared with anchor-based methods \cite{SiamRPN, CascadedRPN, SiamRPN++, SiamDW}, anchor-free tracking methods \cite{Siamfc++, SiamCAR, SiamBAN, Ocean} avoid abundant presets of anchor, thereby significantly reducing model hyperparameters. These methods can achieve more flexible target bounding box regression and state-of-the-art performance. Although it is simple and efficient, the fixed template is hard to faithfully express the target appearance and scale variations.

\textbf{DCF-based Tracking}. MOSSE \cite{MOSSE} is the first attempt to learn the filter coefficients by ridge regression in the Fourier domain. As online tracking methods, DCF-based trackers present better adaptability and generalization to appearance and scale variations. Afterwards, a variety of improved strategies further boost the performance of the DCF-based trackers, such as continuous convolution \cite{CCOT}, dynamic updating of the training set \cite{ECO}, spatial regularization \cite{ASRCF}, temporal smoothing regularization \cite{Xutianyang}. 
CFNet \cite{CFnet} attempts to insert the DCF on the template branch of the siamese network. By treating DCF as a layer in the network, the model can be backpropagated so that the entire network can be trained end-to-end. 
The siamese network enhances the representation of the DCF. Meanwhile, the DCF achieves the online updates of the template. ATOM \cite{ATOM} and DiMP \cite{DiMP} achieve new state-of-the-art tracking performance by combining the online update of DCF and the target localization refinement of the modified IOU-Net \cite{IOUnet}.

\textbf{Segmentation-based Tracking.} CSR-DCF~\cite{CSR-DCF} constructs the target mask via the color histogram of the foreground and background. Then, the boundary effect is suppressed well by adding the mask to the filters. SiamMask \cite{Siammask} extends the light segmentation network to the siamese network-based tracking model, significantly enhancing the target representation with the aid of segmentation loss. Compared with the general VOS methods \cite{STM, VideoMatch, FAVOS, VOSDFB}, SiamMask \cite{Siammask} achieves higher tracking speed owing to utilizing the lightweight segmentation network.
D3S \cite{D3S} replaces the target regression branch in the tracking model with a segmentation network. By combining the accurate localization of DCF and the robustness of the segmentation model to target deformation, D3S achieves new state-of-the-art tracking performance. OceanPlus \cite{Oceanpp} uses the attention retrieval mechanism to obtain the coarse target segmentation mask. The coarse mask filters the background noise in the target feature maps, improving the segmentation accuracy. DMB \cite{DMB} stores the historical appearance and spatial positioning information of the target, providing a rich reference for the current target segmentation. Different from these methods, our segmentation-based tracker develops a dynamic compact memory embedding, which greatly improves the compactness and effectiveness of memory embedding.

\subsection{Memory Embedding for VOT and VOS}
Some works~\cite{DynamicMemoryTracking,STM,VOSDFB} resort to learning effective memory embedding that can provide adequate reference for VOT and VOS tasks.
MemTrack \cite{DynamicMemoryTracking} employs a dynamic memory bank to overcome tracking drift caused by the fixed template. By dynamically storing and reading the tracking results in historical frames as well as fusing with initial target template, the target template can be updated more accurately.
STM \cite{STM} stores dense features and masks of historical frames for current pixel-level spatio-temporal information matching. Dense reference information allows STM to handle appearance changes and occlusions well.
In order to avoid memory redundancy and slow query in excessive memory storage, AFB-URR \cite{VOSDFB} proposes an adaptive feature bank to organize the historical information of the target dynamically. It utilizes the weighted average method to merge similar memories, and learns from the cache replacement strategy to eliminate the memory with the least query frequency.
In this work, we propose a CME, which stores only the target features relevant to the existing memory during the tracking process to alleviate memory redundancy and false retrieval.

\subsection{Deformable Features for Visual Analysis}
The fixed pattern is hard to obtain changeable target representation, especially for non-rigid objects. Staple \cite{Staple} introduces color histogram for visual tracking to resist target deformation. SiamAttn \cite{SiamAttn} develops a deformable siamese attention mechanism to generate the self-attention and cross-attention between the template and search feature. Self-attention is employed to extract contextual information. Cross-attention implicitly updates the target template by aggregating the contextual interdependencies between the template and search region. SiamAttn \cite{SiamAttn} obtains outstanding tracking performance. Deformable DETR \cite{Deformable-DETR} applies multi-scale deformable attention modules to replace the Transformer attention \cite{Transformer} for processing feature maps. The deformable attention acts as a filter for the key elements in all feature map pixels, focusing only on a few sampling positions. With flexibility of the weights, this method achieves excellent performance in detection.
To obtain the complete deformation of the target, the query feature performs an exhaustive search on the target template. Moreover, we apply the segmentation mask of the initial frame as the target posterior probability to enhance the extracted target information.

\section{Proposed Method}

\begin{figure*}[t]
	\centering
	\includegraphics[width=0.92\linewidth]{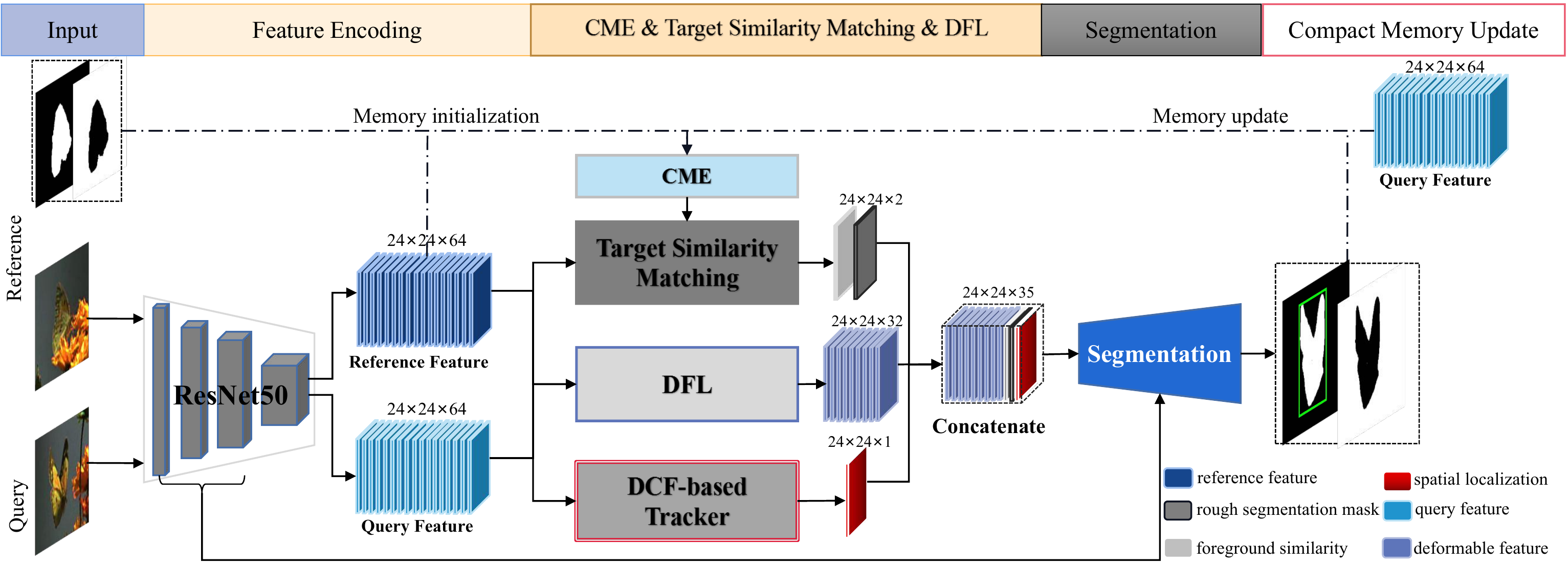}
	\caption{Overall architecture of our method. The query is respectively processed by the DCF-based tracker, DFL module, and target similarity matching equipped with CME. Then, the target foreground similarity, rough segmentation mask, deformable feature, and spatial localization are concatenated together and sent to upscaling segmentation network, outputting the predicted object masks and bounding box.}
	\label{architecture}
	\vspace{-0.3cm}
\end{figure*}

\subsection{Overall Pipeline}
\label{Pipeline}
Fig.~\ref{architecture} outlines the architecture of our method. In order to resist the challenges during the tracking process, our model applies three key components, i.e., target similarity matching based on CME, DFL module, and DCF-based tracker. 
Given the current frame and the initial frame are firstly encoded by the backbone network (i.e., ResNet50 \cite{ResNet}), yielding the searching and reference features.
For computational efficiency, these features are reduced to 64-channels. 
%
Target similarity matching module refers to attention mechanism \cite{Transformer}, which is presented in Section~\ref{Matching}. 
The CME module expands the target feature and segmentation masks to the compact memory, which effectively overcomes the occlusion, similar objects, and appearance variation during tracking.
The DFL module described in Section~\ref{deformable} associates the search feature and the reference feature pixel by pixel, and establishes the correspondence between similar parts of the target . Then the global target correspondence relationship can fully capture the deformation information. The DCF-based tracker is utilized to extract target localization. Referring to the latest deep correlation filters-based tracker, i.e., ATOM \cite{ATOM}, the backbone feature is first reduced to 64 channels by a $1\times1$ convolutional layer. Then, the reduced feature is processed by a $4\times4$ convolution layer and a continuously differentiable activation function (PELU). The 
maximum response position of the activation feature is considered as the target localization. This tracking block is trained by an efficient backprop formulation online (refer to \cite{ATOM} for more details). Finally, the target similarity matching results, the deformable feature, and the target localization are concatenated together along the channel. The combined features are processed by a three stages upscaling segmentation network that is akin to \cite{D3S, Oceanpp}. In each upscaling stage, both the backbone features and the combined features will be first processed by a $3 \times 3$ convolution layer followed by a RELU activation respectively, and the two are subsequently added in an element-wise manner. Then, an up-sampling operation and a $3 \times 3$ convolution layer followed by RELU are performed to obtain a larger spatial resolution.

\subsection{Compact Memory Embedding for Matching}
\label{Matching}

\begin{figure*}[htbp]
	\centering
	\includegraphics[width=0.95\linewidth]{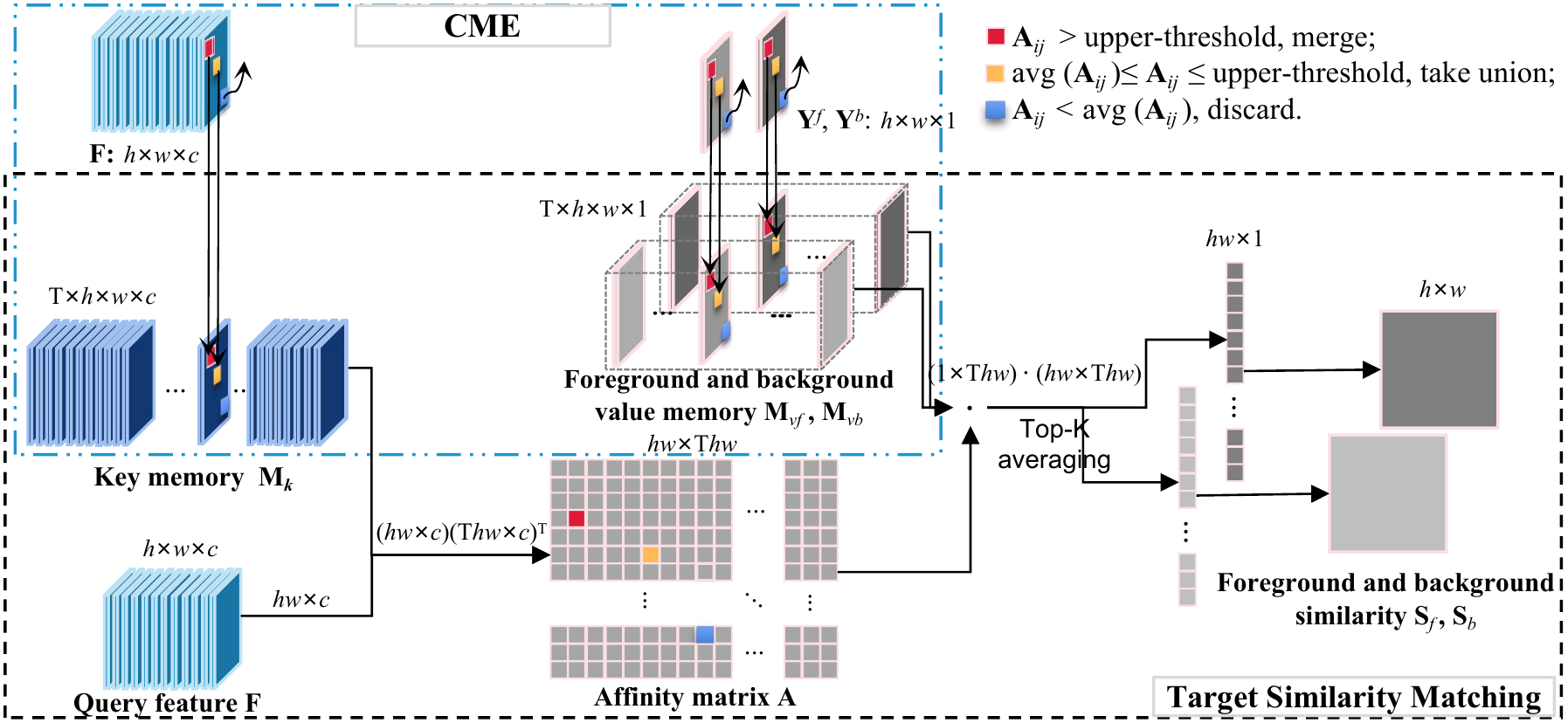}
	\caption{The architecture of the target similarity matching and CME. To capture the target in the current frame, we perform similarity matching between the query feature $\mathbf{F}$ and the target feature memory $\mathbf{M}_{k}$. Specifically, the reshaped $\mathbf{F}$ and $\mathbf{M}_{k}$ construct the affinity matrix $\mathbf{A} \in \mathbb{R}^{hw \times {\rm{T}} hw}$ of the current feature and memory via matrix multiplication. Then, $\mathbf{A}$ retrieves the value memories ($\mathbf{M}_{vf}$ and $\mathbf{M}_{vb}$). The retrieved affinity matrix conducts Top-K averaging along the column dimensions to obtain the foreground and background target similarity $\mathbf{S}_{f}$ and $\mathbf{S}_{b}$. After completing the target segmentation and tracking, $\mathbf{F}$ and the obtained target foreground and background segmentation masks ($\mathbf{Y}^{f}$ and $\mathbf{Y}^{b}$) will be updated to the memory, according to our proposed compact memory adjustment mechanism.}
	\label{Dynamic-Memory}
\vspace{-0.3cm}
\end{figure*}

\textbf{Target Similarity Matching}. Precisely separating the target from the complex background requires effective reference information. Matching-based VOS methods \cite{VideoMatch} make full use of the annotation information in the initial frame to accurately match the current target. In this work, the target similarity matching module consult the conventional attention mechanism \cite{Transformer} which consists of query, key, and value. Fig.~\ref{Dynamic-Memory} shows the structure of the target similarity matching. The query $\mathbf{F}_{t}\in \mathbb{R}^{h \times w \times c}$ is the reduced backbone representation of the current frame. In the common attention mechanism \cite{Transformer}, key and value are different transformations of current frame. Differently, the key $\mathbf{R}\in \mathbb{R}^{h \times w \times c}$ in our model is the backbone feature of the initial frame, and the values ($\mathbf{Y}^{f}_{1}\in \mathbb{R}^{h \times w \times 1}$ and $\mathbf{Y}^{b}_{1}\in \mathbb{R}^{h \times w \times 1}$) are the foreground and background segmentation masks of the first frame.

The target usually undergoes obvious appearance and structure changes in the video sequence. Merely using the fixed initial information of the object cannot guarantee the quality of target  similarity matching. When a new frame, such as $I_{t-1}$, is segmented by the model, the target query $\mathbf{F}_{t-1}$ and the obtained masks ($\mathbf{Y}_{t-1}^{f}$ and $\mathbf{Y}_{t-1}^{b}$) will be taken union with historical information together. The constituted memory embeddings ($\mathbf{M}_{k} \in \mathbb{R}^{{\rm{T}} \times h \times w \times c}$, $\mathbf{M}_{vf} \in \mathbb{R}^{{\rm{T}} \times h \times w \times 1}$, and $\mathbf{M}_{vb} \in \mathbb{R}^{{\rm{T}} \times h \times w \times 1}$) retain rich target appearance information via inserting historical information into the key and value.

In order to establish the pixel-level association of the target between the key memory and query, we first generate the affinity matrix $\mathbf{A}$. Specifically, for better matching, the key memory $\mathbf{M}_{k}$ and query $\mathbf{F}_{t}$ are processed by per-pixel $L_{2}$ normalization along each channel. For convenience, they are still written as ${\mathbf{M}}_{k}$ and ${\mathbf{F}}_{t}$. Moreover, ${\mathbf{M}}_{k}$ and ${\mathbf{F}}_{t}$ are reshaped into the size of ${\rm{T }} h w \times c$ and $h w  \times c$, respectively.
\begin{align}
\mathbf{A}={\mathbf{F}}_{t} * ({\mathbf{M}}_{k})^{T},
\end{align}
where $*$ denotes matrix multiplication; $(\cdot)^{T}$ stands for matrix transpose; $\mathbf{A} \in \mathbb{R}^{(h w) \times ({\rm{T}}h w)}$.

The affinity matrix $\mathbf{A}$ measures the similarity of each pixel between the query map ${\mathbf{F}}_{t}$ and the key memory ${\mathbf{M}}_{k}$. It requires to further retrieve the value memory map to obtain the accurate matching target. Then, the foreground and background value memory maps $\mathbf{M}_{vf}$, and $\mathbf{M}_{vb}$ are reshaped into ${\rm{T}}h w \times 1$, and are expressed as the vectors $\mathbf{m}_{vf}$ and $\mathbf{m}_{vb}$. For ${i \in [1, hw]}$, the affinity vector $\mathbf{a}^{i} \in \mathbb{R}^{{\rm{T }} h w}$ retrieves the value memory vectors $\mathbf{m}_{vf} \in \mathbb{R}^{{\rm{T }} h w \times 1}$ and $\mathbf{m}_{vb}$ via the dot product calculation,
\begin{align}
\hat{\mathbf{s}}_{f}^{i}=\mathbf{a}^{i}\cdot \mathbf{m}_{vf},\\
\hat{\mathbf{s}}_{b}^{i}=\mathbf{a}^{i}\cdot \mathbf{m}_{vb},
\end{align}
where $\hat{\mathbf{s}}_{f}^{i}$ and $\hat{\mathbf{s}}_{b}^{i} \in \mathbb{R}^{{\rm{T }} h w}$.

A matching score with high confidence ensures the accuracy of target matching. Thereby, we apply the top-K averaging function to extract the target score in the retrieved vectors $\hat{\mathbf{s}}_{f}^{i}$ and $\hat{\mathbf{s}}_{b}^{i}$. 
\begin{align}
{\mathbf{s}}_{f}^{i} =\frac{1}{K} \sum_{\hat{\mathbf{s}}_{f}^{i j} \in \operatorname{Top}\left(\hat{\mathbf{s}}_{f}^{i}, K\right)} \hat{\mathbf{s}}_{f}^{i j},\\
{\mathbf{s}}_{b}^{i} =\frac{1}{K} \sum_{\hat{\mathbf{s}}_{b}^{i j} \in \operatorname{Top}\left(\hat{\mathbf{s}}_{b}^{i}, K\right)} \hat{\mathbf{s}}_{b}^{i j},
\end{align}
where the set $\operatorname{Top}(\hat{\mathbf{s}}_{f}^{i}, K)$ denotes the top $K$ matching scores in the vector $\hat{\mathbf{s}}^{i}_{f}$, and $j$ is the subscript index of each score in the first $K$ maximum match scores described above. The finally obtained $\mathbf{s}^{i}_{f}$ is a scalar value. $K$ is set to 3 in our experiments. Target background matching is same as foreground. 
After completing the hw pairs of vector point multiplication and taking the Top-K average operation, the foreground and background similarity vectors $\mathbf{s}_{f} \in \mathbb{R}^{hw}$ and $\mathbf{s}_{b} \in \mathbb{R}^{hw}$ are obtain. Then we transform their dimension, and you can obtain the foreground and background similarity matrices $\mathbf{S}_{f} \in \mathbb{R}^{h \times w}$ and $\mathbf{S}_{b} \in \mathbb{R}^{h \times w}$.
Besides, processing ${\mathbf{S}}_{f}$ and ${\mathbf{S}}_{b}$ by the $softmax(\cdot)$ function generates the rough target posterior probability $\mathbf{P}^{\prime}$, i.e., the rough target foreground segmentation mask.
\begin{align}
\mathbf{P}^{\prime}=softmax\left(\mathbf{S}_{f}, \mathbf{S}_{b}\right).
\end{align}


\label{memory}
\textbf{Compact Memory Embedding}. Abundant memory enable to improve the accuracy of target similarity matching effectively. However, limited by the storage capacity of the computing device, it is impossible to store all historical frame information into the memory, especially for long-term videos.
Besides, since the targets in adjacent frames  tend to be similar, storing such target information will cause redundancy in memory embedding and unnecessary matching queries.
%
Inspired by the Hash~\cite{Monga2006PerceptualIH}, we develop a compact memory adjustment mechanism for our model, forming a diverse and compact target memory. The keys stored in the hash data structure are different from each other. Hash also utilizes keys for information retrieval, which has a higher efficiency of information search. Similarly, we can construct a diverse and non-redundant memory for target similarity matching. Fig.~\ref{Dynamic-Memory} illustrates the structure of the dynamic CME. The affinity matrix $\mathbf{A}$ measures the similarity between the current query feature and the existing memory. Based on $\mathbf{A}$, we merge the high similarity (above the upper threshold) parts between query feature and existing memory. Query feature with moderate similarity to existing memory will be expanded into memory storage. To avoid mismatching caused by low-quality memory, low-similarity target features are directly discarded.

\begin{figure*}[htbp]
\centering
	\begin{minipage}{0.16\linewidth}
		\vspace{2pt}
		\centerline{\includegraphics[width=\textwidth]{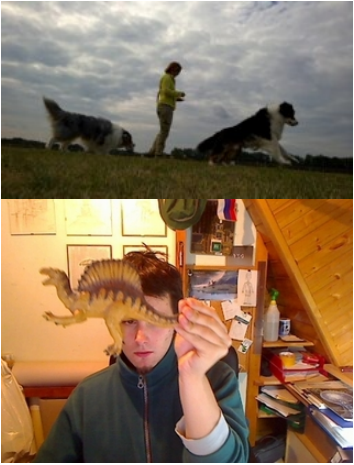}}
		\centerline{Original}
	\end{minipage}
	\begin{minipage}{0.16\linewidth}
		\vspace{2pt}
		\centerline{\includegraphics[width=\textwidth]{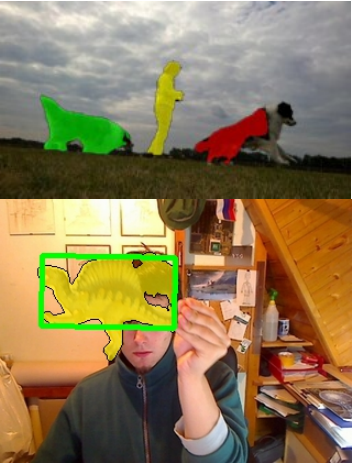}}
	 
		\centerline{(a) No memory}
	\end{minipage}
	\begin{minipage}{0.16\linewidth}
		\vspace{2pt}
		\centerline{\includegraphics[width=\textwidth]{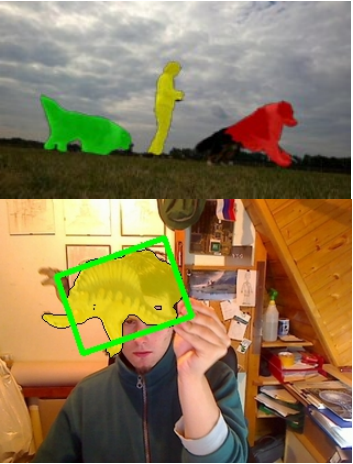}}
	 
		\centerline{(b) All memory}
	\end{minipage}
	\begin{minipage}{0.16\linewidth}
		\vspace{2pt}
		\centerline{\includegraphics[width=\textwidth]{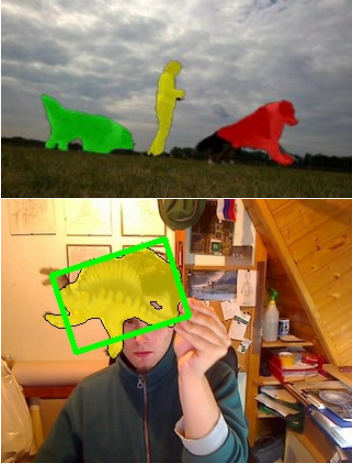}}
	 
		\centerline{(c) AFB \cite{VOSDFB}}
	\end{minipage}
 	\begin{minipage}{0.16\linewidth}
		\vspace{2pt}
		\centerline{\includegraphics[width=\textwidth]{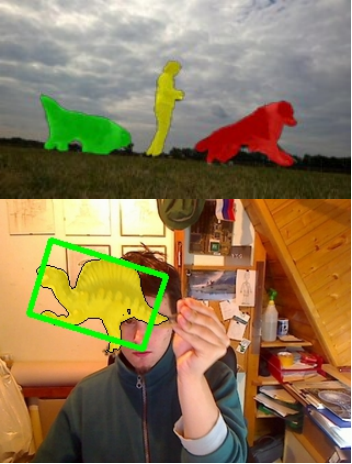}}
	 
		\centerline{(d) CME}
	\end{minipage}
	\caption{The visual comparisons of different memory embeddings. Our CME achieves accurate tracking and segmentation.  On DAVIS2017 \cite{DAVIS17}, the $\mathcal{JF}_{\mathcal{M}}$ score of the four are 60.0, 58.8, 59.4, and 61.2. But improper memory embeddings produce false similarity matching.}
	\label{mask2}
	\vspace{-0.4cm}
\end{figure*}

Most of the VOS methods based on matching \cite{VideoMatch, OSMN} employ the initial frame as a reference template, since the initial feature labeled with ground-truth has an accurate and complete target description. 
Besides, in view of the calculation error of the model itself, 
therefore, we employ the target information of the first frame to initialize the memory embedding and utilize it as the main part of the memory. 
Although the target in the recent video frames is the close to the target of the current frame, 
yet we will reduce the corresponding query preference during the process of target similarity matching. 
During model inference, for example, the video frame $I_{t-1}$ completes the target segmentation and obtains the target query $\mathbf{F}_{t-1}$, as well as the foreground $\mathbf{Y}^{f}_{t-1}$ and background $\mathbf{Y}^{b}_{t-1}$ segmentation masks. In order to extract useful reference information, we first compare $\mathbf{F}_{t-1}$ with the existing memory key $\mathbf{M}_{k}$ to find the similar parts of the two.
The affinity matrix $\mathbf{A} \in \mathbb{R}^{(h w) \times ({\rm{T}}h w)}$ generated in the target similarity matching process measures the correlation between query features $\mathbf{F}_{t-1}$ and key memory $\mathbf{M}_{k}$. Therefore, we directly employ $\mathbf{A}$ for dynamic management of the memory embedding. For each feature element $\mathbf{F}_{t-1}\left(i\right)\in \mathbb{R}^{c} \left(i\in [1, hw]\right)$ in $\mathbf{F}_{t-1}$, we search $\mathbf{A}$ to get the maximum similarity with memory element $\mathbf{M}_{k}\left(j\right)\in \mathbb{R}^{c} \left(j\in [1, \rm{T}hw]\right)$,
\begin{align}
Re\left(\mathbf{F}_{t-1}\left(i\right)\right)=\max _{\forall {j \in [1, \rm{T}hw]} } \mathbf{A}\left(i j\right),
\end{align}
when $\mathbf{A}\left(i j\right)$ takes the maximum value, we denote $j$ as $j^{\prime}$.

If the maximum correlation $\mathbf{A}\left(i j\right)$ between the two is greater than a certain upper limit $\zeta$, we consider the two to be similar enough. In the Hash algorithm~\cite{Monga2006PerceptualIH}, the records in the Hash map with the identical key will be inserted into the same storage space. Therefore, we only save one of multiple similar features into the memory. Considering the diversity of memory, we employ weights to merge over-similar features and corresponding memory. This avoids unnecessary retrievals and memory redundancy. According to the above analysis, the initial reference information is most accurate. Therefore, we employ a small fusion weight $\beta$ to update the current features to the existing memory embedding to avoid the interference of model errors. The online update of the memory embedding is formulated as
\begin{align}
\mathbf{M}_{k}\left(j^{\prime}\right)&= \beta \mathbf{F}_{t-1}\left(i\right) + \left(1-\beta\right)  \mathbf{M}_{k}\left(j^{\prime}\right), \\
\mathbf{M}_{vf} \left(j^{\prime}\right)&=\beta \mathbf{Y}^{f}_{t-1}\left(i\right) + \left(1-\beta\right)  \mathbf{M}_{vf}\left(j^{\prime}\right), \\
\mathbf{M}_{vb} \left(j^{\prime}\right)&=\beta \mathbf{Y}^{b}_{t-1}\left(i\right) + \left(1-\beta\right)  \mathbf{M}_{vb}\left(j^{\prime}\right).
\end{align}

For the maximum correlation $Re\left(\mathbf{F}_{t-1}\left(i\right)\right) < \zeta $, we select features with correlation value higher than the average value 
\begin{align}
{\rm{avg}}\left(Re\left(\mathbf{F}_{t-1}\left(i\right)\right)\right) =  \frac{1}{2e}\left(\frac{1}{wh} \sum_{i \in {wh}} {Re\left(\mathbf{F}_{t-1}\left(i\right)\right)}\right),
\end{align}
where $e$ is Euler's Number. Then we expand them into the existing memory to ensure the diversity of memory. Meanwhile, the query operation with irrelevant memory is avoided, so as to achieve efficient and compact memory storing via the following operations,
\begin{align}
\widetilde{\mathbf{M}}_{k}\left(j^{\prime}\right) &=\operatorname{Union}\left(\mathbf{M}_{k}\left(j^{\prime}\right) , \mathbf{F}_{t-1}\left(i\right)\right), \\
\widetilde{\mathbf{M}}_{vf}\left(j^{\prime}\right) &=\operatorname{Union}\left(\mathbf{M}_{vf}\left(j^{\prime}\right) , \mathbf{Y}^{f}_{t-1}\left(i\right)\right),\\
\widetilde{\mathbf{M}}_{vb}\left(j^{\prime}\right) &=\operatorname{Union}\left(\mathbf{M}_{vb}\left(j^{\prime}\right) , \mathbf{Y}^{b}_{t-1}\left(i\right)\right),
\end{align}
where $\operatorname{Union}(\cdot)$ denotes taking union operation of the current feature and corresponding memory.

Fig.~\ref{mask2} presents the comparisons of our compact memory embedding and two other related methods. As shown in the first row, storing all historical memories and adaptive feature bank (AFB \cite{VOSDFB}) improves the discriminative capability of target similarity matching to a certain extent. But for complex background clutter, redundant memories cause false target matching. Our memory embedding method takes the diversity and compactness of the features, and can achieve much better performance of target similarity matching.

\subsection{Deformable Feature Learning}
\label{deformable}


\begin{figure*}[htbp]

\centering
	\begin{minipage}{0.16\linewidth}
		\vspace{2pt}
		\centerline{\includegraphics[width=\textwidth]{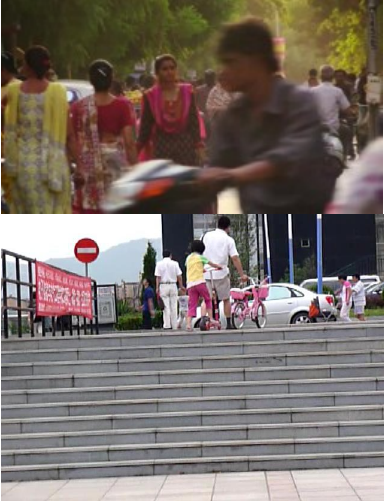}}
		\centerline{Original}
	\end{minipage}
	\begin{minipage}{0.16\linewidth}
		\vspace{2pt}
		\centerline{\includegraphics[width=\textwidth]{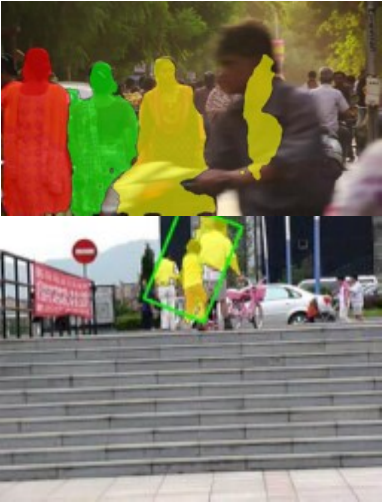}}
		\centerline{(a) Baseline}
	\end{minipage}
	\begin{minipage}{0.16\linewidth}
		\vspace{2pt}
		\centerline{\includegraphics[width=\textwidth]{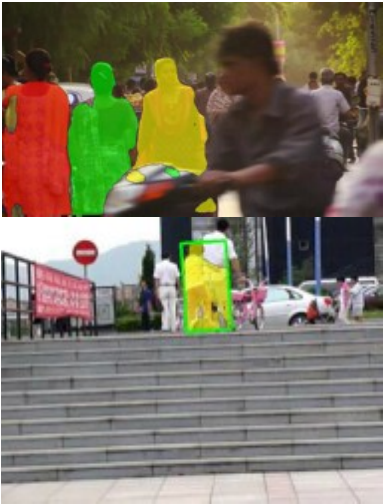}}
		\centerline{(b) +CME}
	\end{minipage}
	\begin{minipage}{0.16\linewidth}
		\vspace{2pt}
		\centerline{\includegraphics[width=\textwidth]{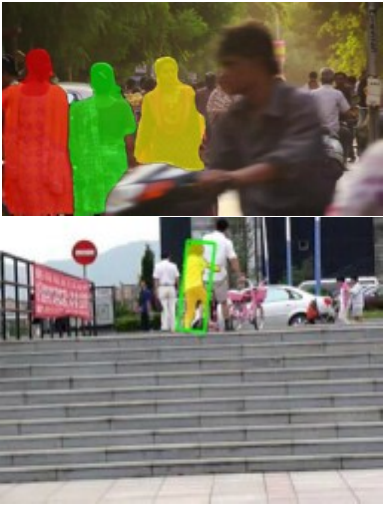}}
		\centerline{(c) +CME+DFL}
	\end{minipage}
 
	\caption{CME effectively improves the discrimination of the model against similar interference. However, the model is unable to segment the edge details of the target completely. The DFL effectively solves this problem and realizes the accurate segmentation of the target.}
	\label{mask3}
\vspace{-0.4cm}
\end{figure*}
As illustrated in Fig.~\ref{mask3}, with the assistance of CME, target similarity matching obtains better discriminative capability that can well solve tracking challenges, such as similar distractors and background interference. 
It also has certain advantages in solving target deformation. But it is unable to solve the severe target deformation or spatial details effectively. Inspired by the graph attention mechanism \cite{GAT}, we propose DFL to further alleviate the aforementioned dilemma. We construct the association between each pixel in the query $\mathbf{F}$ and the whole key $\mathbf{R}$ to capture complete target deformation information. Since the query and the key contain different representation of the target, for per pixel vector $\mathbf{f}^{i}\in \mathbb{R}^{c} \left(i\in[1, wh]\right)$  in the query, we apply the non-shared transformations and calculate the
association of each pixel vector $\mathbf{r}^{j} \in \mathbb{R}^{c} \left(j\in[1, wh]\right)$ in the key to it, yielding the learnable pixel-wise similarity function,
\begin{align}
z_{i j}=\left(W_{F} \mathbf{f}^{i}\right)^{T}\left(W_{R} \mathbf{r}^{j}\right),
\end{align}
where $W_{F}$ and $W_{R}$ indicate the learnable linear transformations which transform $\mathbf{f}^{i}$ and $\mathbf{r}^{j}$ into higher-level representation.

To facilitate the comparison of the similarity between $\mathbf{f}^{i}$ and different parts in $\mathbf{R}$, we normalize $z_{i j}$ across all pixels of $\mathbf{R}$ via the softmax function, yielding the normalized similarity weight,
\begin{align}
d_{i j}= softmax\left(z_{i j}\right)=
\frac{\exp \left(z_{i j}\right)}{\sum_{ \mathbf{r}^{j} \in \mathbf{R}} \exp \left(z_{i j}\right)}.
\end{align}

The normalized pixel-wise similarity weight coefficients $d_{i j}$ are utilized to aggregate the pixel-by-pixel features in $\mathbf{R}$, thus generating the weighted deformation feature corresponding to $\mathbf{f}^{i}$ as
\begin{align}
\mathbf{v}_{i}=\sum_{\mathbf{r}^{j} \in \mathbf{R}} d_{i j} \ \phi_{v} \left(\mathbf{r}^{j}\right),
\end{align}
where $\phi_{v} \left(\cdot \right)$ denotes $\operatorname{ReLU}\left(W_{v}* (\cdot)\right)$ aiming to extract higher-level representation of $\mathbf{r}^{j}$. Then we apply residual connection\cite{ResNet} to cascade the transformed query feature and the weighted deformation feature together, yielding the enhanced feature that contains the deformation information as
\begin{align}
\widetilde{\mathbf{f}}^{i}=\phi_{c} \left(\operatorname{Concat}\left(\mathbf{v}_{i} , \phi_{v} \left(\mathbf{f}^{i}\right) \right)\right),
\end{align}
where $\phi_{c} \left(\cdot \right)$ denotes $\operatorname{ReLU}\left(W_{c}* (\cdot)\right)$ aiming at reducing the dimensionality of features, and $\operatorname{Concat}(\cdot)$ denotes cascade operation.
There are some unavoidable background mismatches in the matched target deformation features. Therefore, we employ the target posterior probability (i.e., the target segmentation mask) $\mathbf{P}$ of the initial frame to enhance the confidence of the deformation feature. Specifically, The obtained target deformation $\widetilde{\mathbf{f}}^{i}$ retrieves the foreground probability $\mathbf{p}^{i}_{1}$ in the initial frame to generate the final deformable feature by dot product,
\begin{align}
\widehat{\mathbf{f}}^{i}=\widetilde{\mathbf{f}}^{i} \cdot \mathbf{p}^{i}_{1}.
\end{align}

\subsection{Tracking with CME and DFL}
This section outlines the application of the proposed CME and DFL to general visual object tracking. 

\textbf{Initialization.} In VOT task, a video sequence merely contains a given bounding box label. We first generate a pseudo mask with the ground-truth box in the initial frame. Then, the pseudo label is utilized to initialize our model and is converted into a more accurate target mask. The converted target mask is employed to initialize the CME module and the template of DFL module.

\textbf{Tracking.} During tracking, we extract an image patch four times the size of the target as the query area at the previous target localization. Then, the query region is processed by our CMEDFL model and obtains the target segmentation mask ${\mathbf{Y}}^{f}_{t}$. Moreover, The sum of each pixel in the background mask ${\mathbf{Y}}^{b}_{t}$ and the corresponding point in ${\mathbf{Y}}^{f}_{t}$ is 1. Meanwhile, the obtained query feature ${\mathbf{F}}_{t}$ and target segmentation mask are used to update the CME in an online manner. Finally, the output of our tracking model is target segmentation mask.
However, most visual object tracking benchmarks take the bounding box as tracking result. Therefore, we firstly binarize the segmentation mask with a threshold of 0.5. Then the target region is extracted on the binary mask using the contour detection function in the OpenCV function library. We employ the bounding box fitting method proposed in VOT\cite{VOT16} benchmark to transform the extracted target region into the target bounding box as the final tracking result.

\section{Experiments}
\subsection{Implementation Details}
In this work, we utilize the first four stages of ResNet50 \cite{ResNet} pre-trained on ImageNet as the backbone network to extract features. 
Object segmentation is a task of pixel-level classification which needs to employ features with high confidence semantics. 
Therefore, we extract the fourth stage of the backbone network for target similarity matching and deformation feature extraction. Then, the backbone features are reduced to 64-channels via $1\times1$ convolution layer followed by a $3\times3$ convolution layer and ReLU activation. In the up-sampling segmentation process, we use the first three stages of the backbone to supplement the target spatial detail information. We set the top-K as $K=3$. The upper-threshold $\zeta$ of the similar memory merging is set to 0.90. 
The fusion weight $\beta$ is set to 0.001.
The above settings are fixed in all related experiments.

\textbf{Network Training}. Both the CME and DCF-based localization modules are updated online without being pre-trained. But, target similarity matching, deformable feature extracting, and up-scaling segmentation network are pre-trained on the Youtube-VOS \cite{YouTubeVOS} dataset. Similar to the sampling strategy utilized in the siamese network-based tracking model, a pair of images with masks are sampled from the video sequence to construct the training samples. We minimize the cross-entropy loss via Adam optimizer \cite{adam} with a learning rate of $8\times10^{-4}$ that has 0.2 decay every 15 epochs. The whole training process takes 60 epochs on an Nvidia Titan XP.

\subsection{Evaluation on Tracking Datasets}
To verify the effectiveness of the proposed model, we conduct extensive evaluations on six popular and challenging tracking benchmarks, including VOT2016 \cite{VOT16}, VOT2018 \cite{VOT18}, VOT2019 \cite{VOT19}, GOT-10k \cite{GOT10K}, TrackingNet \cite{TrackingNet}, and LaSOT\cite{LaSOT}. 
In all comparisons, the best three results are labeled in {\color{red}{red}}, {\color{blue}{blue}}, and {\color{green}{green}} fonts, respectively. Then, we will analyze in detail the experimental results obtained on each dataset.

The VOT datasets~\cite{VOT16,VOT18,VOT19} are currently one of the most convincing and active visual tracking benchmarks. Each dataset contains 60 color video sequences with refined rotated bounding box annotations. VOT updates and supplements the data every year, which puts higher requirements on the performance of the trackers. The official VOT toolkit \cite{VOT16} provides three evaluation criteria, i.e., accuracy (average overlap), robustness (or reliability which is related to failure rate), and the EAO (expected average overlap). EAO comprehensively considers the accuracy and robustness of the trackers. These benchmarks usually rank trackers by EAO score.

\begin{table}[htbp]
\centering
\caption{Comparison results on VOT2016 dataset.}
\label{tab:VOT16}
\setlength{\tabcolsep}{1.4mm}{
\begin{tabular}{ccccc}
    \hline
	Model  	 &Source	&Accuracy $\uparrow$   &Robustness $\downarrow$  &EAO $\uparrow$  \\
    \hline
    SiamFC \cite{Siamfc}	&{ECCVW2016}	&0.530						&0.460				 &0.235\\
	CCOT \cite{CCOT} 	&{ECCV2016}			&0.520                 &0.238                  &0.331\\
	Staple \cite{Staple} 	&{CVPR2016}			&0.544                 &0.378                  &0.295\\
	CSR-DCF \cite{CSR-DCF}	&{CVPR2017}		&0.510			&0.238				&0.338\\
	ECO \cite{ECO} 		&{CVPR2017}		&0.540             &-                  &0.374\\
    SiamRPN \cite{SiamRPN}	&{CVPR2018}	&0.560             &0.302                &0.344\\
    MemTrack \cite{DynamicMemoryTracking} &{ECCV2018} &0.531                &{{0.373}}                  &{{0.272}} \\
    ASRCF \cite{ASRCF}		&{CVPR2019}		&0.560			&0.187			&0.391\\
	SPM \cite{SPM}	 &{CVPR2019}   			&0.620						&0.210				 &0.434\\
	C-RPN \cite{CascadedRPN}    &{CVPR2019}    &0.594                       &-                    &0.363\\
	SiamRPN++ \cite{SiamRPN++}	&{CVPR2019}	&{0.640}      		&0.200                     &0.464\\
	SiamMask \cite{Siammask}   &{CVPR2019} 		&0.639               &0.214                    &0.433\\
	SiamDW\cite{SiamDW} &{CVPR2019}  &0.580  &0.240  &0.370\\
	Update-Net \cite{Updatenet}    &{ICCV2019}      &0.610           	&0.210    &{{0.481}}\\
	ATOM \cite{ATOM}    &{ICCV2019}  &0.610       &{{0.180}}                     &0.430\\
	D3S \cite{D3S}		&{CVPR2020}			&{\color{green}{0.660}}			&{\color{red}{0.131}} &{\color{green}{0.493}}\\
	ROAM++ \cite{ROAM}    &{CVPR2020}  &0.599         &{{0.174}}                     &0.441\\
	SiamAttn \cite{SiamAttn}  &{CVPR2020}  &{\color{red}{0.680}}           &{\color{green}{0.140}}          &{\color{blue}0.537}\\
	CMEDFL	 &ours								&{\color{blue}{0.671}}			&{\color{blue}{0.135}}			&{\color{red}{0.563}}\\
\hline
\end{tabular}
}
\end{table}

\textbf{Quantitative Results on VOT2016}. We compare our model with 18 state-of-the-art trackers on VOT2016 \cite{VOT16} dataset, including the tracking with segmentation methods (CSR-DCF \cite{CSR-DCF}, SiamMask \cite{Siammask}, D3S \cite{D3S}), DCF-based trackers (ASRCF \cite{ASRCF}, CCOT \cite{CCOT}, Staple \cite{Staple}, ATOM \cite{ATOM}), and recent deep trackers (SPM \cite{SPM}, Update-Net \cite{Updatenet}, SiamRPN++ \cite{SiamRPN++}, SiamRPN \cite{SiamRPN}, C-RPN \cite{CascadedRPN}, SiamAttn\cite{SiamAttn}, ROAM++\cite{ROAM}, and MenTrack\cite{DynamicMemoryTracking} ). Besides, CCOT \cite{CCOT} is the best-performing tracker on VOT2016 challenge \cite{VOT16}. Table \ref{tab:VOT16} reports the comparison results. Our model achieves almost the best performance with the excellent EAO score of 0.563 and Accuracy score of 0.671. 
Compared to the second and third best methods SiamAttn\cite{SiamAttn} and D3S\cite{D3S}, our CMEDFL improves EAO score by 2.6$\%$ and 7$\%$, respectively.
Besides, comapred to recent trackers ATOM\cite{ATOM} and ROAM++\cite{ROAM}, CMEDFL obtains the EAO score gain of 13.3$\%$ and 12.2$\%$.
In terms of Accuracy score, SiamAttn\cite{SiamAttn} achieves the best score with 0.680, which surpasses CMEDFL by 0.9$\%$. 
CMEDFL outperforms recent state-of-the-art trackers, i.e., D3S\cite{D3S}, ROAM++\cite{ROAM}, ATOM\cite{ATOM}, SiamRPN++\cite{SiamRPN++}, and SiamMask\cite{Siammask} by 1.1$\%$, 7.2$\%$, 6.1$\%$, 3.1$\%$, 3.2$\%$, respectively. Therefore, the above comparison results prove the effectiveness of the proposed compact memory embedding and deformation feature learning method.

\begin{table}[htbp]
\centering
\caption{Performance comparisons on VOT-2018.}
\label{tab:VOT18}
\setlength{\tabcolsep}{1.2mm}{
\begin{tabular}{ccccc}
    \hline
	Model           &Source				&Accuracy $\uparrow$   &Robustness $\downarrow$  &EAO $\uparrow$  \\
    \hline
    SiamFC \cite{Siamfc}     &{ECCVW2016}     &{0.500}                 &0.590                 &0.188  \\
    CCOT \cite{CCOT}        &{ECCV2016}         &0.494                &0.318                 &0.267\\
    CSR-DCF \cite{CSR-DCF}	&{CVPR2017}		&0.491			&0.356				&0.256\\
	ECO \cite{ECO}         &{CVPR2017}        &0.483                &0.276                 &0.280\\
	SiamRPN \cite{SiamRPN}      &{CVPR2018}     &{0.586}                 &0.276                 &0.83  \\
	MemTrack \cite{DynamicMemoryTracking} &{ECCV2018}  &0.524                &{{0.357}}                  &{{0.248}} \\
	LADCF \cite{LADCF}        &{TIP2019}          &0.503                 &0.159                 &0.389\\
	SPM \cite{SPM}        &{CVPR2019}     &{0.580}                 &0.300                 &0.338  \\
	C-RPN \cite{CascadedRPN}	&{CVPR2019}		&0.550			&0.320				&0.289\\
    SiamRPN++ \cite{SiamRPN++}    &{CVPR2019}   &0.604                &0.234                   &0.417  \\
    SiamMask \cite{Siammask}       &{CVPR2019}      &{{0.609}}                 &0.276                 &0.380  \\
    SiamDW\cite{SiamDW} &{CVPR2019}  &0.520  &0.410  &0.300\\
	Update-Net \cite{Updatenet}    &{ICCV2019}        &-                 				&-                      &{0.393}\\
    ATOM  \cite{ATOM}        &{ICCV2019}        &0.590                &0.204                 &0.401  \\
	DiMP \cite{DiMP}		&{ICCV2019}     &0.597                &{{0.152}}            &0.440\\
	Ocean \cite{Ocean}     &{ECCV2020}      &0.598                &0.169                   &{\color{green}{0.467}}  \\
	D3S \cite{D3S}          &{CVPR2020}  &{\color{blue}{0.640}}                &{\color{green}{0.150}}                  &{\color{blue}{0.489}}\\
	Retina-MAML \cite{Retina-MAML}  &{CVPR2020} &0.604                &0.159                 &0.452 \\
	SiamBAN \cite{SiamBAN}	&{CVPR2020}	&0.597			&0.178			&0.452\\
	ROAM++ \cite{ROAM}   &{CVPR2020}   &0.543                &{{0.195}}                     &0.380\\
	SiamAttn \cite{SiamAttn}  &{CVPR2020}   &{\color{green}{0.630}}           &{0.160}          &0.470\\
	RDTrack \cite{RDTrack}  &{CVPR2021}  &0.595           &{\color{blue}{0.131}}          &0.470\\
	 CMEDFL            &Ours        &{\color{red}{0.641}}        &0.169        &{\color{red}{0.525}}  \\
\hline
\end{tabular}
}
\end{table}

\textbf{Quantitative Results on VOT2018}. On VOT2018 \cite{VOT18}, our method is compared to 22 state-of-the-art trackers, i.e., anchor-based and anchor-free siamese network-based methods (SiamRPN++ \cite{SiamRPN++}, C-RPN \cite{CascadedRPN}, SiamMask \cite{Siammask}, SiamAttn\cite{SiamAttn}, RDTrack\cite{RDTrack}, SiamBAN \cite{SiamBAN} and Ocean \cite{Ocean}), meta-learning based method (Retina-MAML \cite{Retina-MAML}, ROAM++\cite{ROAM}), 
segmentation-based tracking methods (D3S \cite{D3S}, SiamMask \cite{Siammask}), DCF-based trackers (ATOM \cite{ATOM}, DiMP \cite{DiMP}, ECO\cite{ECO}, LADCF \cite{LADCF} which is the champion tracker of VOT2018). As shown in Table~\ref{tab:VOT18}, our method obtains the top-ranking performance with the best EAO of 0.525 and accuracy of 0.641, which exhibits obvious superiority over other methods. Both D3S \cite{D3S} and Ocean-on\cite{Ocean} achieve remarkable performance with the EAO score of 0.489. Compared to recent state-of-the-art trackers, suach as RDTrack\cite{RDTrack}, SiamAttn\cite{SiamAttn}, and ROAM++\cite{ROAM}, our CMEDFL achieves the EAO score gain of 5.5$\%$, 5.5$\%$, and 14.5$\%$, respectively. 
Besides, we find that the segmentation-based trackers (D3S \cite{D3S}, SiamMask \cite{Siammask}, and our CMEDFL) are ahead of other methods in terms of Accuracy score.

\textbf{Quantitative Results on VOT2019}. On VOT2019\cite{VOT19}, our method is compared to 13 state-of-the-art deep trackers related to siamese network-based trackers\cite{DynamicMemoryTracking, SiamRPN++, SiamDW, Siammask, SPM, SiamBAN, Ocean,RDTrack,LightTrack}, meta-learning-based method\cite{Retina-MAML}, segmentation-based tracker\cite{Siammask}, DCF-based tracker\cite{ATOM, DiMP}. As presented in Table~\ref{tab:VOT19}, our method obtains the best performance with 0.368 EAO score, 0.643 Accuracy score, and 0.286 Robustness score, which significantly surpasses other methods. Compared to the second best-performing tracker (LightTrack\cite{LightTrack}) and the third best method (Ocean-on\cite{Ocean}), our model achieves the EAO gain of 1.1$\%$ and 1.8$\%$, respectively. In terms of Accuracy score, our method surpasses ATOM\cite{ATOM}, SiamBAN\cite{SiamBAN}, Ocean-on\cite{Ocean}, RDTrack\cite{RDTrack}, and LightTrack\cite{LightTrack} by 4$\%$, 4.1$\%$, 4.9$\%$, 5$\%$, and 9.1$\%$.

\begin{table}[htbp]
\centering
\caption{Evaluation results on the VOT2019 dataset.The Ocean-on stand for online augmented version of Ocean\cite{Ocean}.}
\label{tab:VOT19}
\setlength{\tabcolsep}{1.4mm}{
\begin{tabular}{ccccc}
    \hline
	Model           &Source			&Accuracy $\uparrow$   &Robustness $\downarrow$  &EAO $\uparrow$  \\
    \hline
    MemTrack \cite{DynamicMemoryTracking} &{ECCV2018}  &0.485                &{{0.587}}                  &{{0.228}} \\
    SPM \cite{SPM}		&{CVPR2019} 		&0.577			&0.507				&0.275\\
    SiamRPN++ \cite{SiamRPN++}     &{CVPR2019}    &0.580                &0.446                   &0.292  \\
    SiamMask \cite{Siammask}     &{CVPR2019}           &0.594                 &0.461                 &0.287  \\
    SiamDW  \cite{SiamDW}     &{CVPR2019}            &0.600                &0.467                   &0.299\\
    ATOM \cite{ATOM}    &{ICCV2019}    &{\color{blue}{0.603}}                &0.411                 &0.301  \\
    DiMP \cite{DiMP}		&{ICCV2019} 			&0.582			&0.371				&0.321\\
    Ocean \cite{Ocean}           &{CVPR2020}         &0.590                &{{0.376}}             &{{0.327}}  \\
	Ocean-on \cite{Ocean}        &{ECCV2020}           &0.594          &{{0.316}}             &{\color{green}{0.350}}  \\
	 Retina-MAML \cite{Retina-MAML} &{CVPR2020}  &0.570                &{{0.366}}           &{{0.313}} \\
	SiamBAN \cite{SiamBAN}	&{CVPR2020} 	&{\color{green}{0.602}}			&0.396			&{{0.327}}\\
	RDTrack \cite{RDTrack}  &{CVPR2021}   &0.593           &{\color{blue}{0.306}}          &0.341\\
	LightTrack \cite{LightTrack}   &{CVPR2021}  &0.552           &{\color{green}{0.310}}       &{\color{blue}0.357}\\
	CMEDFL      &Ours    	&{\color{red}{0.643}}                &{\color{red}{0.286}}                   &{\color{red}{0.368}}  \\
\hline
\end{tabular}
}
\end{table}

\textbf{Quantitative Results on GOT-10k}. GOT-10k\cite{GOT10K} is a large-scale tracking dataset consisting of over 10 thousand sequences and 1.5 million annotations of the axis-aligned bounding box. The trackers are evaluated on the test set consisting of 180 videos via an online server. It employs the average overlap (AO) and success rate (SR) as evaluation criteria. We compare our method to 13 state-of-the-art trackers containing DCF-based trackers\cite{CCOT, Staple, ECO, CFnet, ATOM}, siamese network-based trackers\cite{Siamfc, Siamfc++, SiamRPN, SPM, SiamRPN++, SiamCAR, Siammask}, and meta-learning based method\cite{ROAM}. Table~\ref{tab:GOT10K} reports the comparison results. Our model achieves the best AO score of 59.9, which surpasses the second-best performing tracker SiamCAR\cite{SiamCAR} and the third-best performing tracker ATOM\cite{ATOM} by 3$\%$ and 4.3$\%$. Compared to SiamMask\cite{Siammask}, ROAM++\cite{ROAM}, SiamFC++\cite{Siamfc++}, and SiamRPN++\cite{SiamRPN++}, our method achieves the AO gain of 8.5$\%$, 13.4$\%$,  10.6$\%$, and 8.1$\%$, respectively. Moreover, our CMEDFL also obtains the best SR$_{0.75}$ and SR$_{0.50}$ score with 44.8 and 68.2, respectively.

\begin{table}[htbp]
\centering
\caption{Comparison results on GOT-10k.}
\label{tab:GOT10K}
\setlength{\tabcolsep}{2mm}{
\begin{tabular}{ccccc}
    \hline
	Model       &Source			&SR$_{0.75}$ $\uparrow$       &SR$_{0.50}$ $\uparrow$        &AO $\uparrow$  \\
    \hline
    SiamFC \cite{Siamfc}    &{ECCVW2016}     &9.8              &35.3                  &34.8  \\
    CCOT  \cite{CCOT}      &{ECCV2016}   	&10.7               &32.8                  &32.5  \\
    Staple \cite{Staple}	&{CVPR2016}		&8.9		&23.9			&24.6\\
	ECO \cite{ECO}		&{CVPR2017}		&11.1		&30.9			&31.6\\
	CFNet \cite{CFnet}	&{CVPR2017}	&14.4		&40.4			&37.4\\
	SiamRPN-R18 \cite{SiamRPN}	&{CVPR2018}	&27.0		&58.1		&48.3\\
	SPM \cite{SPM}   &{CVPR2019}      &35.9              &59.3                  &51.3  \\
	 SiamRPN++ \cite{SiamRPN++}   &{CVPR2019}    &32.5              &61.6                  &51.8  \\
    SiamMask \cite{Siammask}  &{CVPR2019}   	&36.6               &58.7                  &51.4  \\
    ATOM  \cite{ATOM}     &{ICCV2019}	&{\color{green}{40.2}}               &{\color{green}{63.4}}  &{\color{green}{55.6}} \\
    SiamFC++ \cite{Siamfc++}    &{AAAI2020}    	&32.3               &57.7                  &49.3  \\
	SiamCAR\cite{SiamCAR}	&{CVPR2020}	&{\color{blue}{41.5}}		&{\color{blue}{67.0}}	&{\color{blue}{56.9}}\\
	 ROAM++ \cite{ROAM}  &{CVPR2020}    &23.6                &{53.2}                     &46.5\\
	 CMEDFL    &Ours   	&{\color{red}{44.8}}       &{\color{red}{68.2}}   &{\color{red}{59.9}} \\
\hline
\end{tabular}
}
\end{table}

\textbf{Quantitative Results on TrackingNet}. The TrackingNet\cite{TrackingNet} is a large-scale dataset aimed at object tracking in the Wild. The testing set consists of 511 video sequences. It utilizes three metrics to evaluate the trackers, i.e., precision, normalized precision (N-precision), and success. We fairly conduct the comparisons with 11 excellent trackers including siamese network-based trackers\cite{Siamfc, SiamGraph, CascadedRPN, Updatenet,Siamfc++},  meta-learning based tracker\cite{ROAM}, and DCF-based trackers\cite{Staple, ECO, CFnet, CSR-DCF, ATOM}. The evaluation results are presented in Table \ref{tab:tn}. CMEDFL obtains the best Precision score with 65.1, and surpasses SiamFC++\cite{Siamfc++}, SiamGraph\cite{SiamGraph}, ATOM\cite{ATOM}, and ROAM++\cite{ROAM} by 0.5$\%$, 1.3$\%$, 0.3$\%$, and 2.8$\%$.
Besides, our method achieves the best Success score of 71.3 and Precision of 65.1, which obtains 0.4 $\%$, 1$\%$, and 4.3$\%$ Success score gain than SiamGraph\cite{SiamGraph}, ATOM\cite{ATOM},  and ROAM++\cite{ROAM} respectively. In our model, both deformable features and compact memory contribute to the outstanding performance.

\begin{table}[htbp]
\centering
\caption{Evaluation results on TrackingNet test set.}
\label{tab:tn}
\setlength{\tabcolsep}{0.8mm}{
\begin{tabular}{ccccc}
    \hline
	Model    &Source 	&Precision $\uparrow$  &N-Precision $\uparrow$  &Success $\uparrow$ \\
    \hline
    SiamFC \cite{Siamfc}	&{ECCVW2016}		&53.3			&66.3						&57.1\\
    Staple \cite{Staple}	&{CVPR2016}			&47.0		&60.3					&52.8 \\
    ECO \cite{ECO}      &{CVPR2017}	 	 &49.2               &61.8    &55.4\\
    CSR-DCF \cite{CSR-DCF}	&{CVPR2017}		&48.0		&62.2					&53.4\\
    CFNet \cite{CFnet}		&{CVPR2017}		&53.3		&65.4					&57.8\\
	C-RPN \cite{CascadedRPN}   &{CVPR2019}   	 &61.9               &74.5              			&66.9 \\
	Update-Net \cite{Updatenet}	&{ICCV2019}		&62.5			&{{75.1}}		 &67.7\\
	 ATOM  \cite{ATOM}     &{ICCV2019}   &{\color{blue}{64.8}}    &{\color{red}{77.1}} 	&{{70.3}}\\
	SiamGraph \cite{SiamGraph}   &{MM2020}       	 &{{63.8}}  &{\color{red}{77.1}} 	 &{\color{green}{70.9}}\\
	SiamFC++\cite{Siamfc++}	&{AAAI2020}		&{\color{green}64.6}	&{\color{green}75.8}					&{\color{blue}71.2}\\
	ROAM++ \cite{ROAM}   &{CVPR2020}   &62.3                &{75.4}                     &67.0\\
	CMEDFL       &Ours        &{\color{red}{65.1}}          &{\color{blue}{76.1}} &{\color{red}{71.3}} \\
\hline
\end{tabular}
}
\end{table}

\begin{figure}[htbp]
\centering
	\begin{minipage}{0.493\linewidth}
		\vspace{3pt}
		\centerline{\includegraphics[width=\textwidth]{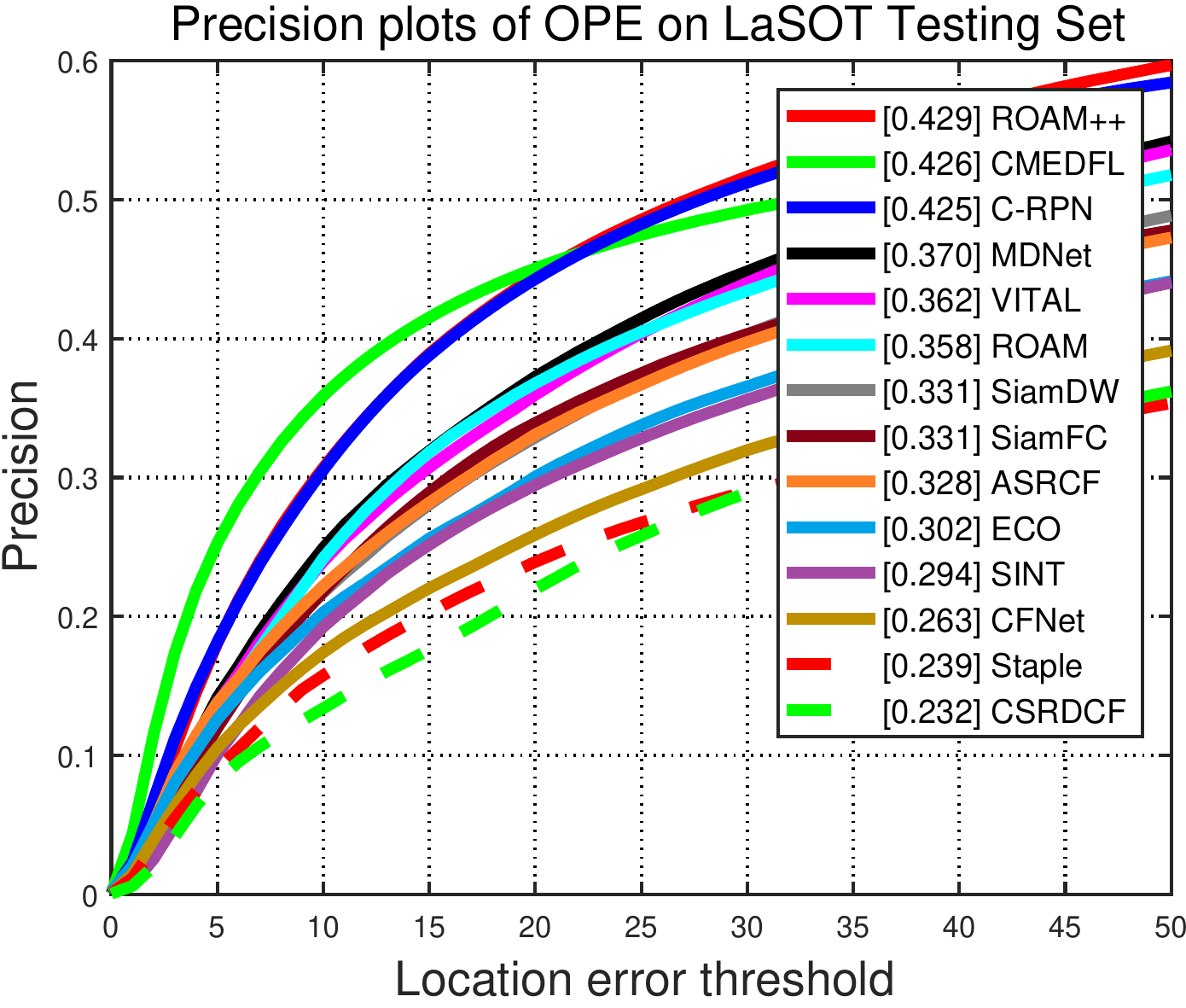}}
		\centerline{(a)Precision plot}
	\end{minipage}
	\begin{minipage}{0.493\linewidth}
		\vspace{3pt}
		\centerline{\includegraphics[width=\textwidth]{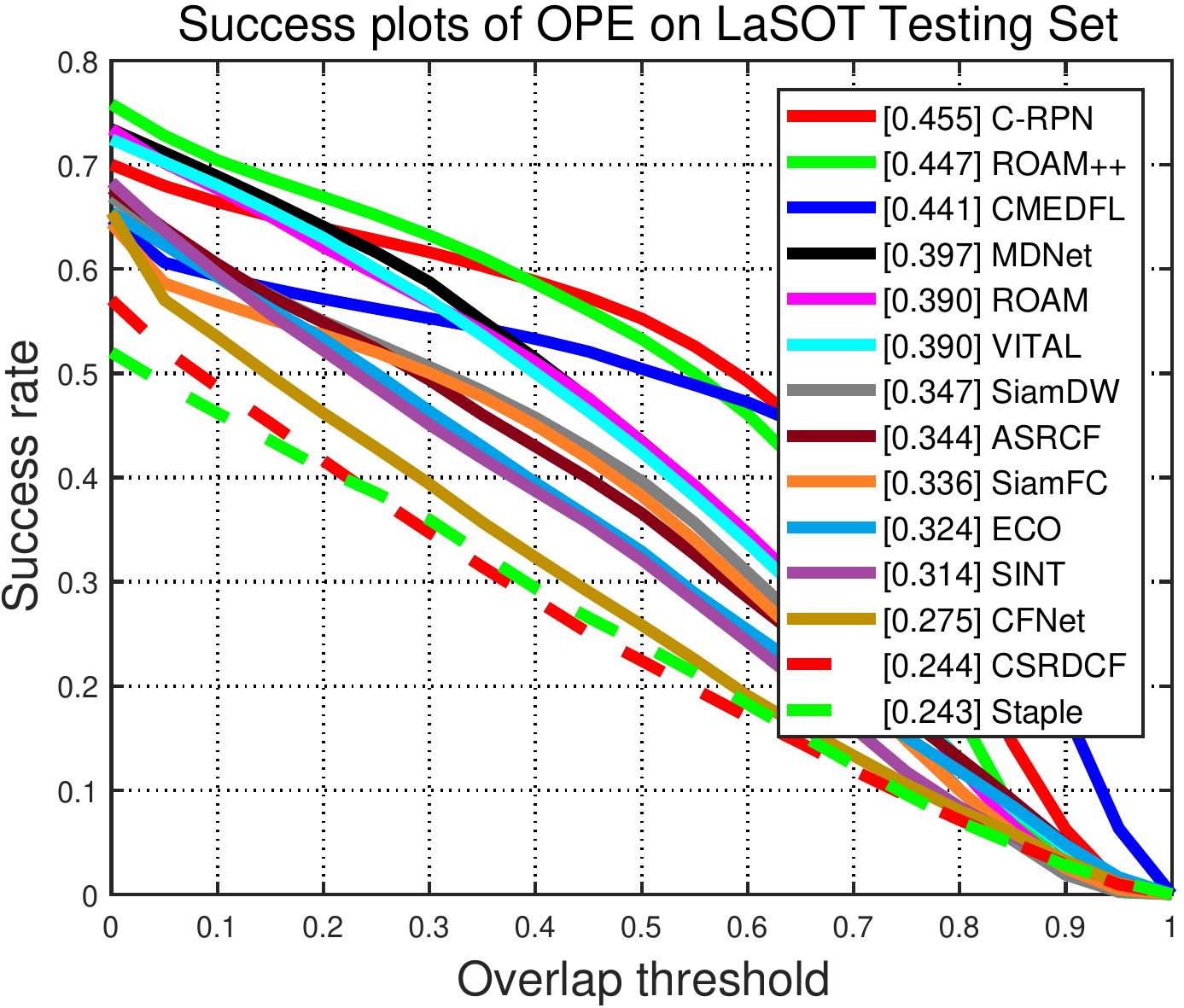}}
		\centerline{(b) Success plot}
	\end{minipage}
 
	\caption{Comparisons with several recent tarcking methods on LaSOT \cite{LaSOT} benchmark in terms of precision and success.}
	\label{lasot}
\end{figure}

\textbf{Quantitative Results on LaSOT}.  LaSOT\cite{LaSOT} is a large-scale visual tracking dataset. The testing set contains 280 video sequences. We conduct the comparisons between our method and 13 state-of-the-art trackers showing in the benchmark\cite{LaSOT}. These methods consist of deep neural network-based trackers (such as SiamFC\cite{Siamfc}, C-RPN\cite{CascadedRPN}, ROAM++\cite{ROAM}, and SiamDW\cite{SiamDW}) and DCF-based tracker\cite{ECO,ASRCF,Staple,CSR-DCF}.  Fig.~\ref{lasot} illustrates the comparison results in terms of precision and success. 
Our CMEDFL tracker obtains 0.426 precision score and 0.441 success score. C-RPN\cite{CascadedRPN} achieves the best success score, and exceeds CMEDFL by 1.4$\%$. However, compared with SiamDW\cite{SiamDW}, ECO\cite{ECO}, and other models, CMEDFL achieves a certain improvement of success score. ROAM++\cite{ROAM} outperforms CMEDFL by 0.3 in terms of Precision score. However,  compared with other trackers, CMEDFL achieves the better precision score.

\begin{figure}[htbp]
	\includegraphics[width=1\linewidth]{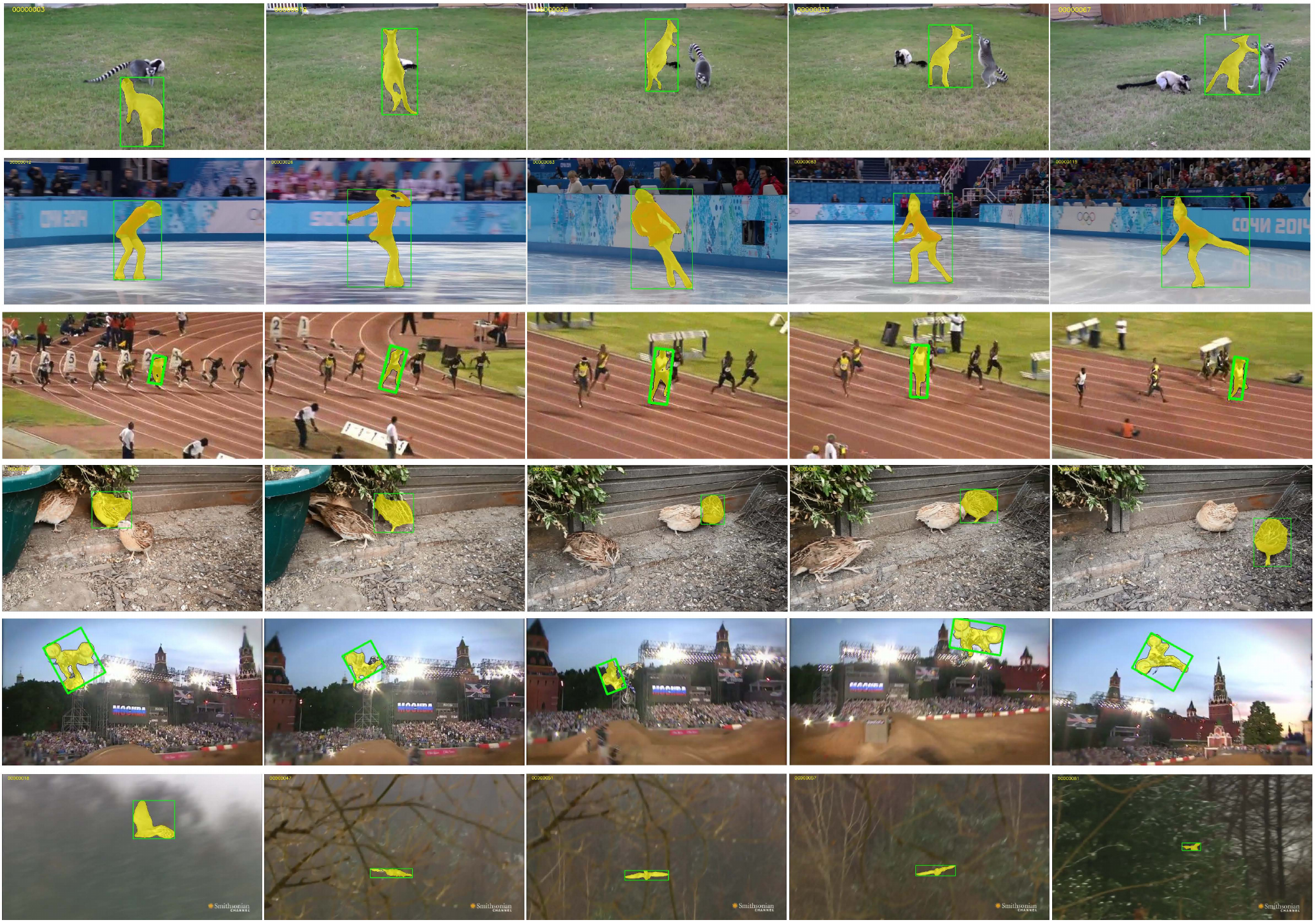}
	\caption{Visualization evaluations of our CMEDFL.}
	\label{tracking}
\end{figure}

\textbf{Qualitative Results}. We conduct visualization experiments on VOT2016\cite{VOT16}, VOT2018\cite{VOT18}, VOT2019\cite{VOT19}, and GOT-10k\cite{GOT10K} datasets.
Fig.~\ref{tracking} illustrates the visualization results of our CMEDFL. These video sequences contain object deformation, similar distractors, background clutter, and other tracking challenges. The first two lines are the deformation case. Although these targets undergo drastic structural changes, our CMEDFL can still segment and track them well. Thus, the effectiveness of the proposed DFL method is verified.
Similar distractor case is illustrated in middle two lines. Similar distractor is one of the inherent challenges in tracking, especially when the target and similar objects block each other, it is easy to cause tracking drift. With the help of CME, our CMEDFL tracker overcomes the obstruction of similar targets successfully. 
The last two lines illustrate the background clutter case. Although the boundary between complex background and target is fuzzy, our method can still segment and track the target well. 
CMEDFL achieves superior segmentation and tracking performance, which verifies that our tracker has certain adaptability and generalization for changes of tracking scene.

\subsection{Evaluation on VOS Task}

 \textbf{Quantitative  Results  on  DAVIS2017}. To further verify the segmentation performance of our model, we conduct evaluations on a popular VOS dataset, i.e., DAVIS2017\cite{DAVIS17}. DAVIS2017\cite{DAVIS17} takes mean Jaccard index ($\mathcal{J}_{\mathcal{M}}$) and mean F-measure ($\mathcal{F}_{\mathcal{M}}$) as measures to evaluate the performance of each model. We compare our method with state-of-the-art VOS methods\cite{OnAVOS, FAVOS, VideoMatch, STM, OSMN} and segmentation-based trackers\cite{D3S, Siammask}. Table \ref{tab:DAVIS17} reports the comparison results. Compared to D3S\cite{D3S} and SiamMask\cite{Siammask}, our tracker achieves the $\mathcal{J}_{\mathcal{M}}$ score gain of 1.9$\%$ and 5.4$\%$, respectively. Although D3S\cite{D3S} surpasses our method by 1.1$\%$ in terms of $\mathcal{F}_{\mathcal{M}}$ score, yet our tracker obtains the best $\mathcal{JF}_{\mathcal{M}}$ score among three segmentation-based trackers. In terms of $\mathcal{JF}_{\mathcal{M}}$ score, CMEDFL outperforms D3S and SiamMask by 0.4$\%$ and 4.8$\%$, respectively.
Even compared with some VOS models, i.e., FAVOS\cite{FAVOS}, VM\cite{VideoMatch}, and OSMN\cite{OSMN}, our method still has superior performance. In terms of inference speed, SiamMask\cite{Siammask} directly utilizes the cross-correlated map as the segmentation input, so it realizes a high processing speed with 55 FPS. Since CMEDFL adopts the memory matching method which contains large calculation, the tracking speed is only 21 FPS. The segmentation-based trackers only apply the light segmentation network. While achieving superior segmentation accuracy, their inference speed is significantly higher than several segmentation methods.

\begin{table}[htbp]
\centering
\caption{Evaluation results on DAVIS2017.}
\label{tab:DAVIS17}
\setlength{\tabcolsep}{1.8mm}{
\begin{tabular}{cccccc}
    \hline
	  Model     &Source		&$\mathcal{J}_{\mathcal{M}}$ $\uparrow$    &$\mathcal{F}_{\mathcal{M}}$ $\uparrow$     &$\mathcal{JF}_{\mathcal{M}}$ $\uparrow$			&FPS $\uparrow$\\
    \hline
	 CMEDFL      &Ours    	&\textbf{59.7}                 &62.7                      &\textbf{61.2}	&21.0 \\
	 D3S \cite{D3S}   &CVPR2020      			&57.8                 &\textbf{63.8}                         &60.8  	&25.0\\
	 SiamMask \cite{Siammask}  &CVPR2019    &54.3                            &58.5                 &56.4 	     &\textbf{55.0} \\
    \hline
    STM \cite{STM}         &ICCV2019		&\textbf{69.2}             &\textbf{74.0}         &\textbf{71.6}	&3.4\\
	 FAVOS \cite{FAVOS}     &CVPR2018 		&54.6                    &61.8                    &58.2     &0.8\\
	OSMN \cite{OSMN}      &CVPR2018		& 52.5                    &57.1                          &54.8		&\textbf{8.0}\\
	VM \cite{VideoMatch}    &ECCV2018 	&56.6                 &-        &-  	&3.1 \\
	OnAVOS \cite{OnAVOS}    &BMVC2017   	&61.6             &69.1                     &65.4		&0.1 \\
\hline
\end{tabular}
}
\end{table}

\begin{figure}[htbp]
	\centering
	\includegraphics[width=1\linewidth]{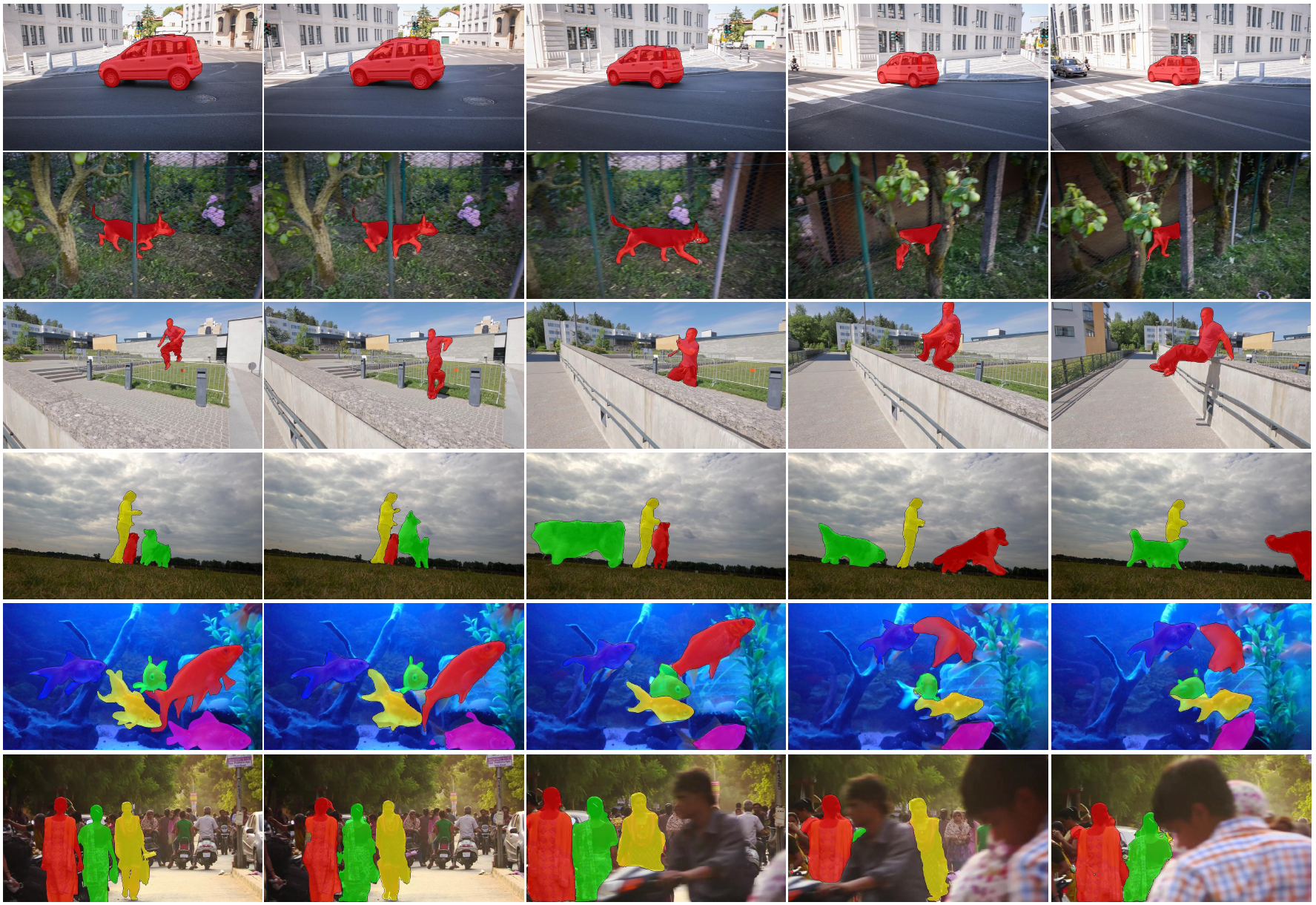}
	\caption{Visual evaluations of our method on DAVIS2017. The sequences contain various challenges (occlusion, deformation, etc.). The target objects are marked in red, green, yellow, and others.}
	\label{fig:SMVOS}
\vspace{-0.4cm}
\end{figure}

\textbf{Qualitative Results}. Fig.~\ref{fig:SMVOS} illustrates the qualitative results of our tracker on DAVIS2017. Even if the object undergoes occlusion, deformation, and similar distractors, our method still achieves accurate target segmentation.

\begin{table*}[htbp]
\caption{Component-wise comparisons of the proposed method on VOT2016, VOT2018, and VOT2019 benchmarks.}\label{tab:Ablation}
\centering
\setlength{\tabcolsep}{2.9mm}{
\begin{tabular}{cccc|ccc|ccc}
\hline
 \multirow{2}{*}{Methods}  &\multicolumn{3}{c|}{VOT2016} & \multicolumn{3}{c|}{VOT2018}  &\multicolumn{3}{c}{VOT2019} \\ \cline{2-10}
    &{Accuracy}&{Robustness}&{EAO}  &{Accuracy}&{Robustness}&{EAO} &{Accuracy}&{Robustness}&{EAO}\\
\hline
{Baseline}&{{0.642}}&{0.135}&{0.480} &{0.639}&{0.159}&{0.473}&{0.632}&{0.321}&{0.301}\\
{$+\rm{DFL}_{N}$}&{0.655}&{0.154}&{0.498} &{0.633}&{0.159}&{0.482}&{0.631}&{0.321}&{0.309}\\
{$\rm{+DFL}$}&{{0.657}}&{\textbf{0.112}}&{0.520}&\textbf{0.641} &\textbf{0.150}&{{0.496}} &{0.640} &{0.336}&{{0.313}} \\
{$+\rm{ME}_{ALL}$}&{0.656}&{0.163}&{0.517} &{0.619}&{0.187}&{0.488}&{0.619}&{0.341}&{0.313}\\
{$+\rm{ME}_{AFB}$}&{0.637}&{0.154}&{0.516}&{0.612}&{0.183}&{0.498}&{0.606}&{0.351}&{0.318}\\
{$+\rm{CME}$}&{{0.666}}&{{0.149}} &{0.525}     &{0.638}&{0.187}&{{0.504}} &{0.637} &{0.356}&{{0.317}}\\
{$+\rm{CME+DFL}_{N}$}&{0.669}&{0.135}&{0.541}&{0.632}&{0.150}&{0.506}&{0.636}&{0.301}&{0.333}\\
{$+\rm{ME}_{ALL}+\rm{DFL}$}&{0.667}&{0.158}&{0.524} &{0.634}&{0.183}&{0.505}&{0.629}&{0.316}&{0.344}\\
{$+\rm{ME}_{AFB}+\rm{DFL}$}&{0.661}&{0.154}&{0.523} &{0.619}&{0.187}&{0.500}&{0.607}&{0.296}&{0.360}\\
{$\rm{+CME+DFL}$}&{\textbf{0.671}}&{{0.135}} &\textbf{0.563}  &{\textbf{0.641}}&{0.169}&{\textbf{0.525}} &\textbf{0.643} &\textbf{0.286}&{\textbf{0.368}}\\
\hline
\end{tabular}
}
\end{table*}

\subsection{Ablation Study}
\textbf{Component-wise Comparison.} To verify the efficacy of the proposed method, we conduct an ablation study of the proposed modules on VOT2016~\cite{VOT16}, VOT2018~\cite{VOT18}, and VOT2019~\cite{VOT19}. The experimental results are listed in Table~\ref{tab:Ablation}. The Baseline model contains a backbone network (as introduced in Section~\ref{Pipeline}) and target similarity matching module without memory embedding (as introduced in Section~\ref{Matching}). 
$\rm{DFL}_{N}$ represents the target deformation feature without using the initial frame target mask for post-processing.
DFL is the target deformation feature processed by the target mask of the initial frame.
$\rm{ME}_{ALL}$ indicates that we store the target information in all historical frames as memory.
$\rm{ME}_{AFB}$ is the adaptive memory embedding \cite{VOSDFB}.
CME represents the proposed dynamic compact memory embedding.

Baseline obtains the EAO scores of 0.480, 0.473, and 0.301 on VOT2016, VOT2018, and VOT2019, respectively. Besides, the achieved Accuracy scores are 0.642, 0.639, and 0.632.
With the help of the deformable feature learning module ($\rm{+DFL}$), the EAO score of Baseline increases to 0.520, 0.496, and 0.313, achieving obvious gain of 4$\%$, 2.3$\%$, and 1.2 $\%$, respectively. 
 This verifies that the deformable feature coding enables to promote more robust tracking performance.
 In addition, although the $\rm{+DFL}_{N}$ model improves the tracking performance of Baseline to a certain extent. Specifically, on VOT2016, VOT2018, and VOT2019, an increase of 1.9$\%$, 0.9$\%$, and 0.8$\%$ are achieved respectively. Yet, compared to $\rm{+DFL}$, there are the gap of 2.2$\%$, 1.4$\%$, and 0.4$\%$ EAO scores on VOT2016, VOT2018, and VOT2019, respectively.
 
 Compared with a single target template, memory can provide richer information reference for target similarity matching. Therefore, above three types of memory embedding, $\rm{+ME}_{ALL}$, $\rm{+ME}_{AFB}$ and $+\rm{CME}$, achieves certain performance improvement compared to Baseline. For example, on the VOT2016 benchmark, the above three memory embeddings obtain 3.7$\%$, 3.6$\%$ and 4.5$\%$ of EAO scores improvement respectively compared to Baseline. On VOT2018, the EAO score gains of 1.5$\%$, 2.5$\%$ and 3.1$\%$ are achieved respectively.
The compact memory removes redundant historical reference information, ensuring the credibility of target similarity matching. The $+\rm{CME}$ model significantly improves the EAO score of the Baseline on VOT2016, VOT2018 and VOT2019 by 2.2$\%$, 3.1$\%$ and 1.6$\%$. In terms of EAO and Accuracy scores, $+\rm{CME}$ shows certain performance advantages compared to $\rm{+ME}_{ALL}$ and $\rm{+ME}_{AFB}$. For example, on VOT2018\cite{VOT18}, $+\rm{CME}$ exceeds $\rm{+ME}_{ALL}$ and $\rm{+ME}_{AFB}$ 1.6$\%$ and 0.6$\%$ EAO scores, and achieves 1.9$\%$ and 2.6$\%$ Accuracy scores improvement, respectively. 

Finally, we combine the well-performing $+\rm{CME}$ module and the two deformation feature learning modules mentioned above, as well as experimental comparisons.
With the help of compact memory embedding CME, the EAO scores of $+\rm{DFL}$ and $\rm{+DFL}_{N}$ are improved by 4.3$\%$ on VOT2016, by 2.4$\%$ and 5.5$\%$ on VOT2019,respectively. Therefore, the effectiveness of CME is further verified.
Besides, we evaluate the combinations of the $+\rm{DFL}$ module and above three memory embeddings.
It can be found that the deformation feature learning module $+\rm{DFL}$ can promote the tracking performance of the above three memory embedding models. For example, on VOT2019 data set, $\rm{+ME}_{ALL}+\rm{DFL}$, $\rm{+ME}_{AFB}+\rm{DFL}$ and $\rm{+CME+DFL}$ achieve 3.1$\%$, 4.2$\%$ and 5.1$\%$ EAO scores gains compared to $\rm{+ME}_{ALL}$, $\rm{+ME}_{AFB}$ and $\rm{+CME}$, respectively. 
Moreover, on the VOT2018 data set, the EAO score gains of 1.7$\%$, 0.2$\%$ and 2.1$\%$ are also achieved respectively. Therefore, the effectiveness of the deformation feature learning module DFL proposed in the article is also verified.

\begin{figure}[htbp]
	\centering
	\includegraphics[width=0.92\linewidth]{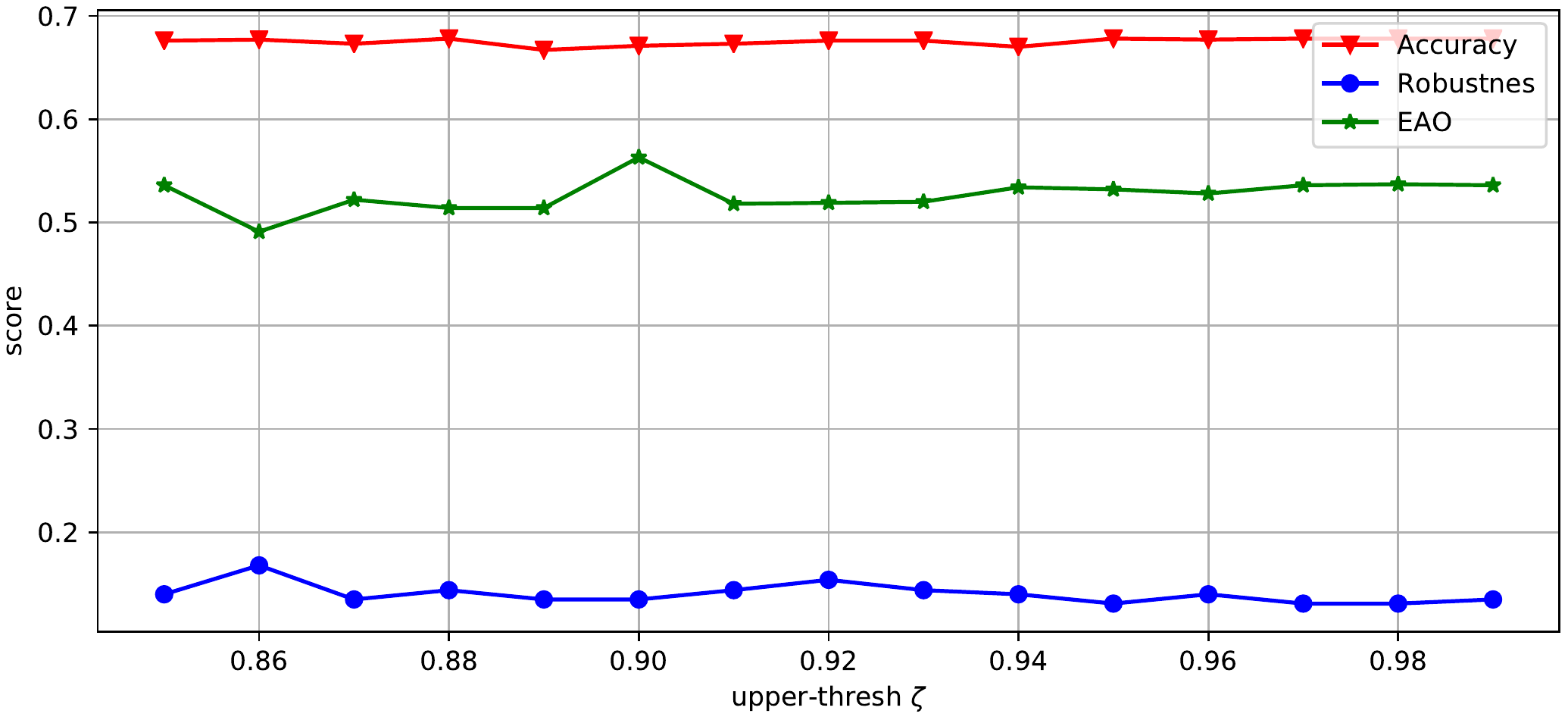}
	\caption{On the VOT2016\cite{VOT16} benchmark, the impact of upper-threshold $\zeta$ in our CMEDFL.}
	\label{fig:param1}
\end{figure}

\begin{figure}[htbp]
	\centering
	\includegraphics[width=0.92\linewidth]{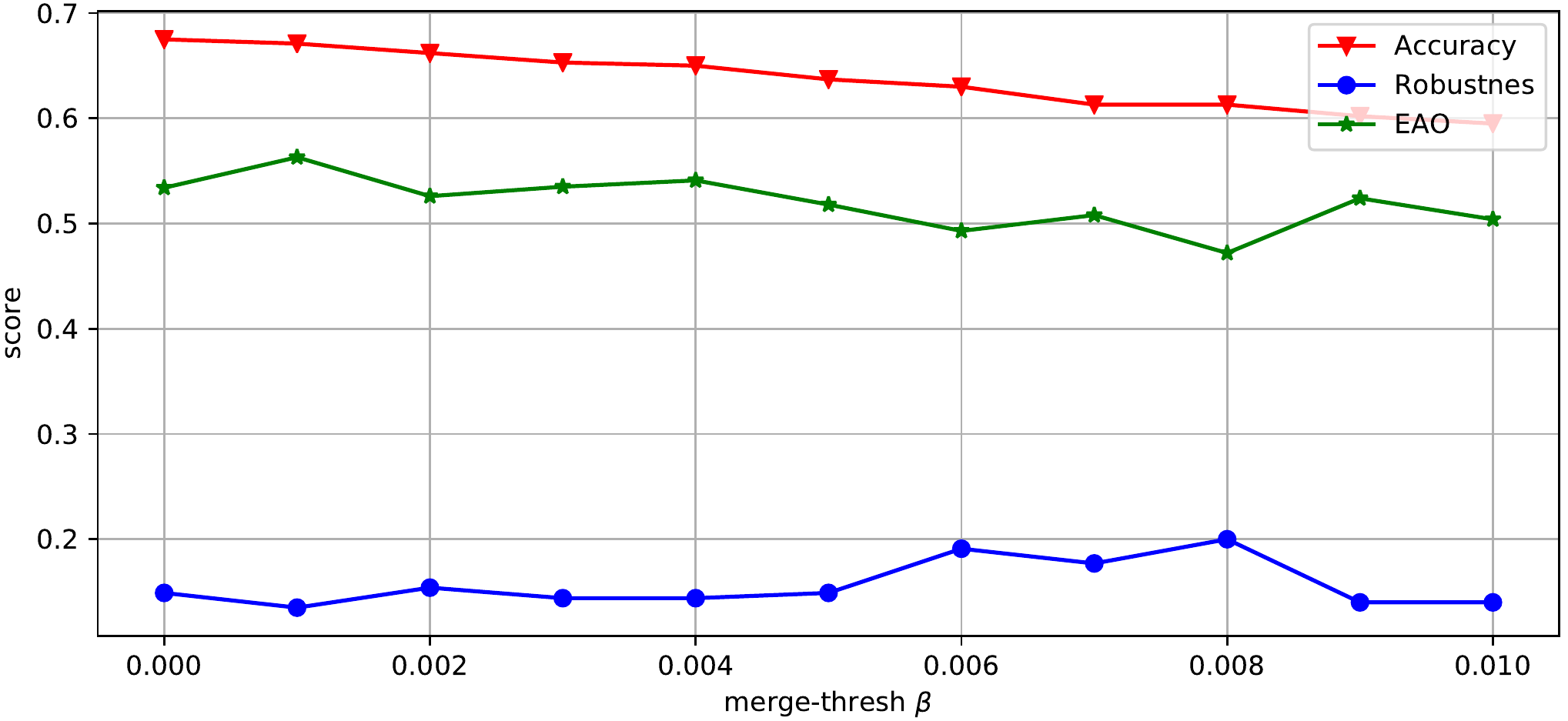}
	\caption{On the VOT2016\cite{VOT16} benchmark, the effect of fusion weight $\beta$ in our CMEDFL.}
	\label{fig:param2}
\end{figure}

\textbf{Influence of Parameters.}
In the section ~\ref{memory}, when the similarity value between the current target feature and the existing memory is higher than the upper-threshold $\zeta$, we think the two are similar enough to be the same. Therefore, the parameter $\zeta$ measures the degree of similarity between the current target feature and the existing memory. 
In order to reduce the redundancy of the memory, we utilize a fusion weight $\beta$ to merge the highly similar parts between the current target feature and the existing memory.
We fix $\beta$ and set it as 0.001. Then, we explore the effect of $\zeta$ on our CMEDFL on VOT2016\cite{VOT16} data set. 
As shown in Fig.~\ref {fig:param1}, the Accuracy score of CMEDFL is relatively stable for the setting of $\zeta$.
But for EAO score, the tracking performance of CMEDFL fluctuates slightly, and when $\zeta$ is set to 0.90, the best EAO score is achieved.
Besides, we fix $\zeta$ and set it as 0.90. 
Fig.~\ref {fig:param2} illustrate the effects of $\beta$ on CMEDFL on the VOT2016\cite{VOT16} data set. As $\beta$ increases, the Accuracy and EAO score of CMEDFL decrease. When $\beta$ is set to 0.001, CMEDFL achieves the best EAO score.

\begin{figure}[htbp]
\centering
	\includegraphics[width=0.8\linewidth]{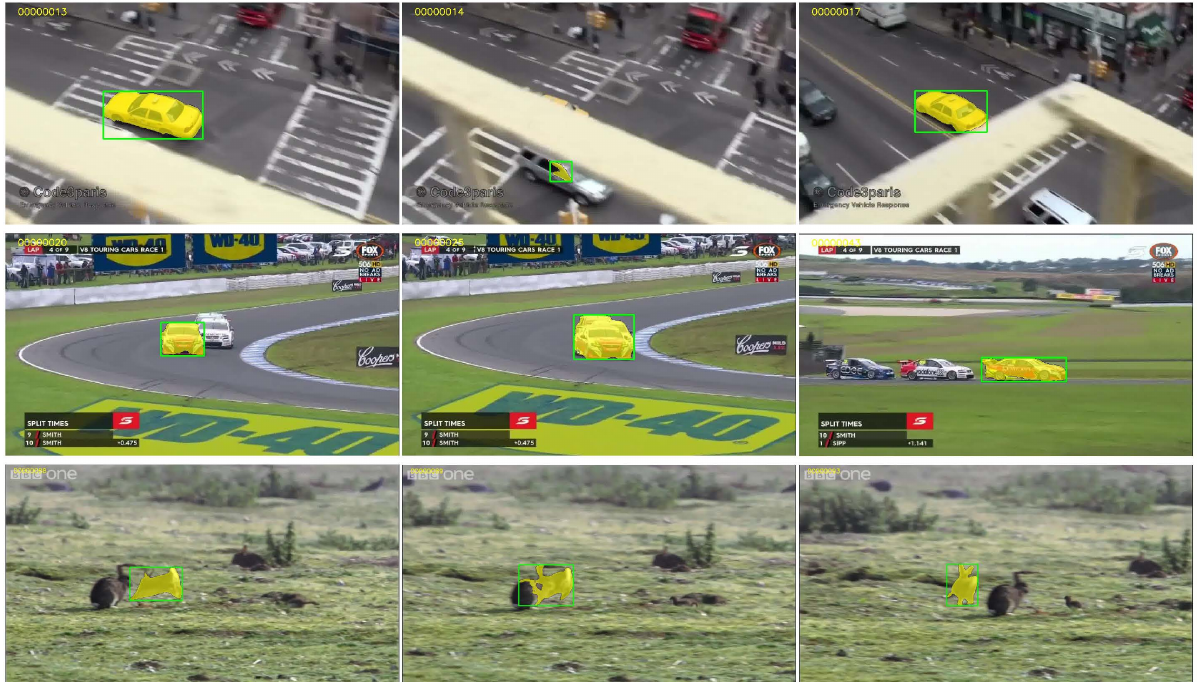}
	\caption{Visualization of failing and recovering cases on GOT-10k \cite{GOT10K}. The mutual occlusion of nearby similar targets will interfere with our tracking model, but when the target is restored to the field of view, our CMEDFL can track the target well again. }
	\label{fail}
\end{figure}

\textbf{Failing and recovering cases.} The overlapping occlusion of adjacent similar targets in a short time can easily lead to tracking failure. Especially for the tracking model updated online, it is easy to cause unreliable model update and even model corruption. This challenge has high requirements for the discrimination of the trackers. 
Fig.~\ref{fail} illustrates the tracking performance of our CMEDFL under this case. We can find that, when the target is occluded, similar object close to the target will cause great interference to the tracking model, which makes the model make wrong judgment. We introduce dynamic compact memory into the model. Moreover, the memory stores the feature information of the target in the historical frames, so it can provide rich information reference for target tracking to avoid tracking drift. Although the target undergoes tracking error in a short time when it is blocked, our model can still track correctly once the target is restored to the field of view.

\textbf{Inference speed}.
On DAVIS2017\cite{DAVIS17} dataset, our trackers achieves the inference speed of 21 fps. On GOT-10k\cite{GOT10K}, TrackingNet\cite{TrackingNet}, and VOT\cite{VOT16,VOT18, VOT19} benchmarks, the tracking speed of CMEDFL tracker is 19 fps.
\section{Conclusions}
In this paper, we propose to learn dynamic CME and deformable feature for deformable visual object tracking.
The CME takes the initial target feature as the basis. We maintain the compactness and diversification of memory by the idea of Hash map. It absorbs the features related to the existing memory in the tracking process, thus avoiding the target mismatch caused by uncorrelated memory and improving the discrimination of the model effectively.
The DFL module establishes a weighted correlation between the pixels-wise query feature and the entire template. The captured target deformation complements detailed information for target segmentation. The outstanding performance achieved on seven challenging benchmarks including VOT and VOS tasks all verify the effectiveness of our model.


\bibliographystyle{IEEEtran}
\bibliography{reference}

\begin{thebibliography}{10}
\providecommand{\url}[1]{#1}
\csname url@samestyle\endcsname
\providecommand{\newblock}{\relax}
\providecommand{\bibinfo}[2]{#2}
\providecommand{\BIBentrySTDinterwordspacing}{\spaceskip=0pt\relax}
\providecommand{\BIBentryALTinterwordstretchfactor}{4}
\providecommand{\BIBentryALTinterwordspacing}{\spaceskip=\fontdimen2\font plus
\BIBentryALTinterwordstretchfactor\fontdimen3\font minus
  \fontdimen4\font\relax}
\providecommand{\BIBforeignlanguage}[2]{{%
\expandafter\ifx\csname l@#1\endcsname\relax
\typeout{** WARNING: IEEEtran.bst: No hyphenation pattern has been}%
\typeout{** loaded for the language `#1'. Using the pattern for}%
\typeout{** the default language instead.}%
\else
\language=\csname l@#1\endcsname
\fi
#2}}
\providecommand{\BIBdecl}{\relax}
\BIBdecl

\bibitem{Xing2010MultipleHT}
J.~Xing, H.~Ai, and S.~Lao, ``Multiple human tracking based on multi-view
  upper-body detection and discriminative learning,'' in \emph{Proc. IEEE Int.
  Conf. Pattern Recognit. (ICPR)}, 2010, pp. 1698--1701.

\bibitem{Liu2012HandPR}
L.~Liu, J.~Xing, H.~Ai, and X.~Ruan, ``Hand posture recognition using finger
  geometric feature,'' in \emph{Proc. IEEE Int. Conf. Pattern Recognit.
  (ICPR)}, 2012, pp. 565--568.

\bibitem{Lee2015OnRoadPT}
K.-H. Lee and J.~Hwang, ``On-road pedestrian tracking across multiple driving
  recorders,'' \emph{IEEE Trans. Multimedia}, vol.~17, pp. 1429--1438, 2015.

\bibitem{Tang2017MultiplePT}
S.~Tang, M.~Andriluka, B.~Andres, and B.~Schiele, ``Multiple people tracking by
  lifted multicut and person re-identification,'' in \emph{Proc. IEEE Conf.
  Comput. Vis. Pattern Recognit. (CVPR)}, 2017, pp. 3701--3710.

\bibitem{Siamfc}
L.~Bertinetto, J.~Valmadre, J.~F. Henriques, A.~Vedaldi, and P.~H.~S. Torr,
  ``Fully-convolutional siamese networks for object tracking,'' in \emph{Proc.
  Eur. Conf. Comput. Vis. Workshop (ECCVW)}, vol. 9914, 2016, pp. 850--865.

\bibitem{SINT}
R.~Tao, E.~Gavves, and A.~W.~M. Smeulders, ``Siamese instance search for
  tracking,'' in \emph{Proc. IEEE Conf. Comput. Vis. Pattern Recognit. (CVPR)},
  2016, pp. 1420--1429.

\bibitem{SiamRPN}
B.~Li, J.~Yan, W.~Wu, Z.~Zhu, and X.~Hu, ``High performance visual tracking
  with siamese region proposal network,'' in \emph{Proc. IEEE Conf. Comput.
  Vis. Pattern Recognit. (CVPR)}, 2018, pp. 8971--8980.

\bibitem{CascadedRPN}
H.~Fan and H.~Ling, ``Siamese cascaded region proposal networks for real-time
  visual tracking,'' in \emph{Proc. IEEE Conf. Comput. Vis. Pattern Recognit.
  (CVPR)}, 2019, pp. 7952--7961.

\bibitem{SiamRPN++}
B.~Li, W.~Wu, Q.~Wang, F.~Zhang, J.~Xing, and J.~Yan, ``Siamrpn++: Evolution of
  siamese visual tracking with very deep networks,'' in \emph{Proc. IEEE Conf.
  Comput. Vis. Pattern Recognit. (CVPR)}, 2019, pp. 4282--4291.

\bibitem{SiamDW}
Z.~Zhang and H.~Peng, ``Deeper and wider siamese networks for real-time visual
  tracking,'' in \emph{Proc. IEEE Conf. Comput. Vis. Pattern Recognit. (CVPR)},
  2019, pp. 4591--4600.

\bibitem{SPM}
G.~Wang, C.~Luo, Z.~Xiong, and W.~Zeng, ``Spm-tracker: Series-parallel matching
  for real-time visual object tracking,'' in \emph{Proc. IEEE Conf. Comput.
  Vis. Pattern Recognit. (CVPR)}, 2019, pp. 3643--3652.

\bibitem{Siamfc++}
Y.~Xu, Z.~Wang, Z.~Li, Y.~Ye, and G.~Yu, ``Siamfc++: Towards robust and
  accurate visual tracking with target estimation guidelines,'' in \emph{Proc.
  Conf. Artif. Intell. (AAAI)}, 2020, pp. 12\,549--12\,556.

\bibitem{SiamCAR}
D.~Guo, J.~Wang, Y.~Cui, Z.~Wang, and S.~Chen, ``Siamcar: Siamese fully
  convolutional classification and regression for visual tracking,'' in
  \emph{Proc. IEEE Conf. Comput. Vis. Pattern Recognit. (CVPR)}, 2020, pp.
  6268--6276.

\bibitem{SiamBAN}
Z.~Chen, B.~Zhong, G.~Li, S.~Zhang, and R.~Ji, ``Siamese box adaptive network
  for visual tracking,'' in \emph{Proc. IEEE Conf. Comput. Vis. Pattern
  Recognit. (CVPR)}, 2020, pp. 6667--6676.

\bibitem{Ocean}
Z.~Zhang, H.~Peng, J.~Fu, B.~Li, and W.~Hu, ``Ocean: Object-aware anchor-free
  tracking,'' in \emph{Proc. Eur. Conf. Comput. Vis. (ECCV)}, vol. 12366, 2020,
  pp. 771--787.

\bibitem{MOSSE}
D.~S. Bolme, J.~R. Beveridge, B.~A. Draper, and Y.~M. Lui, ``Visual object
  tracking using adaptive correlation filters,'' in \emph{Proc. IEEE Conf.
  Comput. Vis. Pattern Recognit. (CVPR)}, 2010, pp. 2544--2550.

\bibitem{ECO}
M.~Danelljan, G.~Bhat, F.~S. Khan, and M.~Felsberg, ``{ECO:} efficient
  convolution operators for tracking,'' in \emph{Proc. IEEE Conf. Comput. Vis.
  Pattern Recognit. (CVPR)}, 2017, pp. 6931--6939.

\bibitem{CCOT}
M.~Danelljan, A.~Robinson, F.~S. Khan, and M.~Felsberg, ``Beyond correlation
  filters: Learning continuous convolution operators for visual tracking,'' in
  \emph{Proc. Eur. Conf. Comput. Vis. (ECCV)}, vol. 9909, 2016, pp. 472--488.

\bibitem{Xutianyang}
T.~Xu, Z.~Feng, X.~Wu, and J.~Kittler, ``Learning adaptive discriminative
  correlation filters via temporal consistency preserving spatial feature
  selection for robust visual object tracking,'' \emph{IEEE Trans. Image
  Process.}, vol.~28, no.~11, pp. 5596--5609, 2019.

\bibitem{ATOM}
M.~Danelljan, G.~Bhat, F.~S. Khan, and M.~Felsberg, ``{ATOM:} accurate tracking
  by overlap maximization,'' in \emph{Proc. IEEE Conf. Comput. Vis. Pattern
  Recognit. (CVPR)}, 2019, pp. 4660--4669.

\bibitem{ASRCF}
K.~Dai, D.~Wang, H.~Lu, C.~Sun, and J.~Li, ``Visual tracking via adaptive
  spatially-regularized correlation filters,'' in \emph{Proc. IEEE Conf.
  Comput. Vis. Pattern Recognit. (CVPR)}, 2019, pp. 4670--4679.

\bibitem{LADCF}
T.~Xu, Z.~Feng, X.~Wu, and J.~Kittler, ``Learning adaptive discriminative
  correlation filters via temporal consistency preserving spatial feature
  selection for robust visual object tracking,'' \emph{IEEE Trans. Image
  Process.}, vol.~28, no.~11, pp. 5596--5609, 2019.

\bibitem{DiMP}
G.~Bhat, M.~Danelljan, L.~V. Gool, and R.~Timofte, ``Learning discriminative
  model prediction for tracking,'' in \emph{Proc. IEEE Int. Conf. Comput. Vis.
  (ICCV)}, 2019, pp. 6181--6190.

\bibitem{Siammask}
Q.~Wang, L.~Zhang, L.~Bertinetto, W.~Hu, and P.~H.~S. Torr, ``Fast online
  object tracking and segmentation: {A} unifying approach,'' in \emph{Proc.
  IEEE Conf. Comput. Vis. Pattern Recognit. (CVPR)}, 2019, pp. 1328--1338.

\bibitem{D3S}
A.~Lukezic, J.~Matas, and M.~Kristan, ``{D3S} - {A} discriminative single shot
  segmentation tracker,'' in \emph{Proc. IEEE Conf. Comput. Vis. Pattern
  Recognit. (CVPR)}, 2020, pp. 7131--7140.

\bibitem{Monga2006PerceptualIH}
V.~Monga and B.~Evans, ``Perceptual image hashing via feature points:
  Performance evaluation and tradeoffs,'' \emph{IEEE Trans. Image Process.},
  vol.~15, pp. 3452--3465, 2006.

\bibitem{VOT16}
M.~Kristan, A.~Leonardis, J.~Matas, M.~Felsberg, R.~P. Pflugfelder, L.~Cehovin,
  T.~Voj{\'{\i}}r, and et~al, ``The visual object tracking {VOT2016} challenge
  results,'' in \emph{Proc. Eur. Conf. Comput. Vis. Workshop (ECCVW)}, vol.
  9914, 2016, pp. 777--823.

\bibitem{VOT18}
M.~Kristan, A.~Leonardis, J.~Matas, M.~Felsberg, R.~P. Pflugfelder, L.~C. Zajc,
  T.~Voj{\'{\i}}r, and et~al, ``The sixth visual object tracking {VOT2018}
  challenge results,'' in \emph{Proc. Eur. Conf. Comput. Vis. Workshop
  (ECCVW)}, vol. 11129, 2018, pp. 3--53.

\bibitem{VOT19}
M.~Kristan, A.~Berg, L.~Zheng, L.~Rout, L.~V. Gool, L.~Bertinetto,
  M.~Danelljan, and et~al, ``The seventh visual object tracking {VOT2019}
  challenge results,'' in \emph{Proc. IEEE Int. Conf. Comput. Vis. Workshop
  (ICCVW)}, 2019, pp. 2206--2241.

\bibitem{GOT10K}
L.~Huang, X.~Zhao, and K.~Huang, ``Got-10k: {A} large high-diversity benchmark
  for generic object tracking in the wild,'' \emph{IEEE Trans. Pattern Anal.
  Mach. Intell.}, vol.~43, no.~5, pp. 1562--1577, 2021.

\bibitem{TrackingNet}
M.~M{\"{u}}ller, A.~Bibi, S.~Giancola, S.~Al{-}Subaihi, and B.~Ghanem,
  ``Trackingnet: {A} large-scale dataset and benchmark for object tracking in
  the wild,'' in \emph{Proc. Eur. Conf. Comput. Vis. (ECCV)}, vol. 11205, 2018,
  pp. 310--327.

\bibitem{LaSOT}
H.~Fan, L.~Lin, F.~Yang, P.~Chu, G.~Deng, S.~Yu, H.~Bai, Y.~Xu, C.~Liao, and
  H.~Ling, ``Lasot: {A} high-quality benchmark for large-scale single object
  tracking,'' in \emph{Proc. IEEE Conf. Comput. Vis. Pattern Recognit. (CVPR)},
  2019, pp. 5374--5383.

\bibitem{VideoMatch}
Y.~Hu, J.~Huang, and A.~G. Schwing, ``Videomatch: Matching based video object
  segmentation,'' in \emph{Proc. Eur. Conf. Comput. Vis. (ECCV)}, vol. 11212,
  2018, pp. 56--73.

\bibitem{FAVOS}
J.~Cheng, Y.~Tsai, W.~Hung, S.~Wang, and M.~Yang, ``Fast and accurate online
  video object segmentation via tracking parts,'' in \emph{Proc. IEEE Conf.
  Comput. Vis. Pattern Recognit. (CVPR)}, 2018, pp. 7415--7424.

\bibitem{DAVIS17}
\BIBentryALTinterwordspacing
J.~Pont{-}Tuset, F.~Perazzi, S.~Caelles, P.~Arbelaez, A.~Sorkine{-}Hornung, and
  L.~V. Gool, ``The 2017 {DAVIS} challenge on video object segmentation,''
  \emph{arXiv:1704.00675}, 2017. [Online]. Available:
  \url{http://arxiv.org/abs/1704.00675}
\BIBentrySTDinterwordspacing

\bibitem{ResNet}
K.~He, X.~Zhang, S.~Ren, and J.~Sun, ``Deep residual learning for image
  recognition,'' in \emph{Proc. IEEE Conf. Comput. Vis. Pattern Recognit.
  (CVPR)}, 2016, pp. 770--778.

\bibitem{CFnet}
J.~Valmadre, L.~Bertinetto, J.~F. Henriques, A.~Vedaldi, and P.~H.~S. Torr,
  ``End-to-end representation learning for correlation filter based tracking,''
  in \emph{Proc. IEEE Conf. Comput. Vis. Pattern Recognit. (CVPR)}, 2017, pp.
  5000--5008.

\bibitem{IOUnet}
B.~Jiang, R.~Luo, J.~Mao, T.~Xiao, and Y.~Jiang, ``Acquisition of localization
  confidence for accurate object detection,'' in \emph{Proc. Eur. Conf. Comput.
  Vis. (ECCV)}, vol. 11218, 2018, pp. 816--832.

\bibitem{CSR-DCF}
A.~Lukezic, T.~Vojir, L.~C. Zajc, J.~Matas, and M.~Kristan, ``Discriminative
  correlation filter with channel and spatial reliability,'' in \emph{Proc.
  IEEE Conf. Comput. Vis. Pattern Recognit. (CVPR)}, 2017, pp. 4847--4856.

\bibitem{STM}
S.~W. Oh, J.~Lee, N.~Xu, and S.~J. Kim, ``Video object segmentation using
  space-time memory networks,'' in \emph{Proc. IEEE Int. Conf. Comput. Vis.
  (ICCV)}, 2019, pp. 9225--9234.

\bibitem{VOSDFB}
Y.~Liang, X.~Li, N.~H. Jafari, and J.~Chen, ``Video object segmentation with
  adaptive feature bank and uncertain-region refinement,'' in \emph{Annual
  Conference on Neural Information Processing Systems. (NeurIPS)}, 2020.

\bibitem{Oceanpp}
\BIBentryALTinterwordspacing
Z.~Zhang, B.~Li, W.~Hu, and H.~Peng, ``Towards accurate pixel-wise object
  tracking by attention retrieval,'' \emph{arXiv:2008.02745}, 2020. [Online].
  Available: \url{https://arxiv.org/abs/2008.02745}
\BIBentrySTDinterwordspacing

\bibitem{DMB}
\BIBentryALTinterwordspacing
F.~Xie, W.~Yang, B.~Liu, K.~Zhang, W.~Xue, and W.~Zuo, ``Discriminative
  segmentation tracking using dual memory banks,'' \emph{arXiv:2009.09669},
  2020. [Online]. Available: \url{https://arxiv.org/abs/2009.09669}
\BIBentrySTDinterwordspacing

\bibitem{DynamicMemoryTracking}
T.~Yang and A.~B. Chan, ``Learning dynamic memory networks for object
  tracking,'' in \emph{Proc. Eur. Conf. Comput. Vis. (ECCV)}, vol. 11213, 2018,
  pp. 153--169.

\bibitem{Staple}
L.~Bertinetto, J.~Valmadre, S.~Golodetz, O.~Miksik, and P.~H.~S. Torr,
  ``Staple: Complementary learners for real-time tracking,'' in \emph{Proc.
  IEEE Conf. Comput. Vis. Pattern Recognit. (CVPR)}, 2016, pp. 1401--1409.

\bibitem{SiamAttn}
Y.~Yu, Y.~Xiong, W.~Huang, and M.~R. Scott, ``Deformable siamese attention
  networks for visual object tracking,'' in \emph{Proc. IEEE Conf. Comput. Vis.
  Pattern Recognit. (CVPR)}, 2020, pp. 6727--6736.

\bibitem{Deformable-DETR}
\BIBentryALTinterwordspacing
X.~Zhu, W.~Su, L.~Lu, B.~Li, X.~Wang, and J.~Dai, ``Deformable {DETR:}
  deformable transformers for end-to-end object detection,''
  \emph{arXiv:2010.04159}, 2020. [Online]. Available:
  \url{https://arxiv.org/abs/2010.04159}
\BIBentrySTDinterwordspacing

\bibitem{Transformer}
A.~Vaswani, N.~Shazeer, N.~Parmar, J.~Uszkoreit, L.~Jones, A.~N. Gomez,
  L.~Kaiser, and I.~Polosukhin, ``Attention is all you need,'' in \emph{Annual
  Conference on Neural Information Processing Systems. (NeurIPS)}, 2017, pp.
  5998--6008.

\bibitem{OSMN}
L.~Yang, Y.~Wang, X.~Xiong, J.~Yang, and A.~K. Katsaggelos, ``Efficient video
  object segmentation via network modulation,'' in \emph{Proc. IEEE Conf.
  Comput. Vis. Pattern Recognit. (CVPR)}, 2018, pp. 6499--6507.

\bibitem{GAT}
P.~Velickovic, G.~Cucurull, A.~Casanova, A.~Romero, P.~Li{\`{o}}, and
  Y.~Bengio, ``Graph attention networks,'' in \emph{Proc. Int. Conf. Learn.
  Represent. (ICLR)}, 2018.

\bibitem{YouTubeVOS}
\BIBentryALTinterwordspacing
N.~Xu, L.~Yang, Y.~Fan, D.~Yue, Y.~Liang, J.~Yang, and T.~S. Huang,
  ``Youtube-vos: {A} large-scale video object segmentation benchmark,''
  \emph{arXiv:1809.03327}, 2018. [Online]. Available:
  \url{http://arxiv.org/abs/1809.03327}
\BIBentrySTDinterwordspacing

\bibitem{adam}
D.~P. Kingma and J.~Ba, ``Adam: {A} method for stochastic optimization,'' in
  \emph{Proc. Int. Conf. Learn. Represent. (LCLR)}, Y.~Bengio and Y.~LeCun,
  Eds., 2015.

\bibitem{Updatenet}
L.~Zhang, A.~Gonzalez{-}Garcia, J.~van~de Weijer, M.~Danelljan, and F.~S. Khan,
  ``Learning the model update for siamese trackers,'' in \emph{Proc. IEEE Int.
  Conf. Comput. Vis. (ICCV)}, 2019, pp. 4009--4018.

\bibitem{ROAM}
T.~Yang, P.~Xu, R.~Hu, H.~Chai, and A.~B. Chan, ``{ROAM:} recurrently
  optimizing tracking model,'' in \emph{Proc. IEEE Conf. Comput. Vis. Pattern
  Recognit. (CVPR)}, 2020, pp. 6717--6726.

\bibitem{Retina-MAML}
G.~Wang, C.~Luo, X.~Sun, Z.~Xiong, and W.~Zeng, ``Tracking by instance
  detection: {A} meta-learning approach,'' in \emph{Proc. IEEE Conf. Comput.
  Vis. Pattern Recognit. (CVPR)}, 2020, pp. 6287--6296.

\bibitem{RDTrack}
S.~Cheng, B.~Zhong, G.~Li, X.~Liu, Z.~Tang, X.~Li, and J.~Wang, ``Learning to
  filter: Siamese relation network for robust tracking,'' in \emph{Proc. IEEE
  Conf. Comput. Vis. Pattern Recognit. (CVPR)}, 2021, pp. 4421--4431.

\bibitem{LightTrack}
B.~Yan, H.~Peng, K.~Wu, D.~Wang, J.~Fu, and H.~Lu, ``Lighttrack: Finding
  lightweight neural networks for object tracking via one-shot architecture
  search,'' in \emph{Proc. IEEE Conf. Comput. Vis. Pattern Recognit. (CVPR)},
  2021, pp. 15\,180--15\,189.

\bibitem{SiamGraph}
F.~Zhao, T.~Zhang, C.~Ma, M.~Tang, J.~Wang, and X.~Wang, ``Siamese attentive
  graph tracking,'' in \emph{Proc. ACM Int. Conf. Multimedia. (MM)}, 2020, pp.
  1542--1550.

\bibitem{OnAVOS}
P.~Voigtlaender and B.~Leibe, ``Online adaptation of convolutional neural
  networks for video object segmentation,'' in \emph{British Machine Vision
  Conference. (BMVC)}, 2017.

\end{thebibliography}

\vspace{-10mm}

\end{document}